\documentclass[11pt]{article}

\usepackage[preprint]{acl}

\usepackage{times}
\usepackage{latexsym}
\usepackage{amsmath}

\usepackage[T1]{fontenc}

\usepackage[utf8]{inputenc}

\usepackage{microtype}

\usepackage{inconsolata}

\usepackage{graphicx}
\usepackage{listings}
\lstdefinelanguage{SQL}{
  morekeywords={SELECT,FROM,WHERE,JOIN,ON,AS,AND,OR,CASE,WHEN,THEN,ELSE,END},
  sensitive=false,
  morecomment=[l]{--},
  morestring=[b]',
}
\title{SpectraQuery: A Hybrid Retrieval-Augmented Conversational Assistant for Battery Science}

\author{
  \textbf{Sreya Vangara\textsuperscript{1}} \quad
  \textbf{Jagjit Nanda\textsuperscript{2}} \quad
  \textbf{Yan-Kai Tzeng\textsuperscript{2}} \quad
  \textbf{Eric Darve\textsuperscript{1,3}} \\
  \\
  \textsuperscript{1}Mechanical Engineering, Stanford University \\
  \textsuperscript{2}Applied Energy Division, SLAC National Accelerator Laboratory \\
  \textsuperscript{3}Institute of Computational and Mathematical Engineering, Stanford University \\
  \\
  \small{
    \textbf{Correspondence:} \href{mailto:svangara@stanford.edu}{svangara@stanford.edu}, \href{mailto:darve@stanford.edu}{darve@stanford.edu}
  }
}

\begin{document}
\maketitle
\begin{abstract}
Scientific reasoning increasingly requires linking structured experimental data with the unstructured literature that explains it, yet most large language model (LLM) assistants cannot reason jointly across these modalities. We introduce \textbf{SpectraQuery}, a hybrid natural-language query framework that integrates a relational Raman spectroscopy database with a vector-indexed scientific literature corpus using a Structured and Unstructured Query Language (SUQL)–inspired design. By combining semantic parsing with retrieval-augmented generation, SpectraQuery translates open-ended questions into coordinated SQL and literature retrieval operations, producing cited answers that unify numerical evidence with mechanistic explanation. Across SQL correctness, answer groundedness, retrieval effectiveness, and expert evaluation, SpectraQuery demonstrates strong performance: approximately \textbf{80\%} of generated SQL queries are fully correct, synthesized answers reach \textbf{93–97\%} groundedness with 10–15 retrieved passages, and battery scientists rate responses highly across accuracy, relevance, grounding, and clarity (\textbf{4.1–4.6/5}). These results show that hybrid retrieval architectures can meaningfully support scientific workflows by bridging data and discourse for high-volume experimental datasets.
\end{abstract}

\section{Introduction}
\subsection{Background and Motivation}
Large Language Models (LLMs) have revolutionized natural-language understanding and reasoning, driving rapid progress in information retrieval, dialogue systems, and autonomous research assistants \cite{zhao_potential_2024}. However, advances in battery science increasingly depend on correlating large volumes of quantitative measurements with qualitative mechanistic insights buried in the literature. For example, operando spectroscopy, electrochemical impedance measurements, and optical microscopy generate rich structured datasets describing evolving electrode materials and cell performance \cite{xue_solutions_2024}. 3D operando Raman spectroscopy in particular is an emerging and promising technique which provides real-time tracking of structural and chemical transformations during charge-discharge cycles \cite{zhu_application_2018}. A visualization of what Raman spectroscopy captures for battery materials is demonstrated in Figure 1. Characteristic peaks such as the A1g mode of transition-metal-oxide vibrations and the D/G carbon bands reflect changes in redox state and structural disorder \cite{ferrari_interpretation_2000}. Each experiment yields tens of thousands of spectra across spatial and temporal coordinates, and recent initiatives in battery science have also begun establishing open and FAIR Raman databases to increase access to this valuable data \cite{coca-lopez_open_2025}.

Unfortunately, though, drawing chemical meaning from these spectroscopic features typically demands cross-referencing dozens of prior literature that describe oxygen redox reactions, side product formation, or crystalline disordering, phase changes, and more \cite{heber_monitoring_2021}. The result is an intensive manual workflow that separates numerical observation from conceptual interpretation. Current LLM–based tools rarely integrate both structured measurement data and unstructured text so that a researcher can pose a single query combining the two. As a result, the workflow remains divided: analysts query numerical datasets in isolation and separately review literature for context, a process that is time-consuming, error-prone, and may overlook cross-modal patterns.

\subsection{Contributions of Work}
SpectraQuery unifies structured experimental data and unstructured text by providing an end-to-end system that performs structured database reasoning, semantic text retrieval, and grounded synthesis in response to natural language questions. Inspired by the SUQL framework of Liu et al. (2023) and built upon the retrieval-augmented generation paradigm of Lewis et al. (2020), the system demonstrates that hybrid querying and generative reasoning can be applied effectively to real scientific datasets \cite{liu_suql_2023, lewis_retrieval-augmented_2020}. By adapting hybrid query architectures originally developed for general-purpose knowledge bases, SpectraQuery extends structured-unstructured retrieval into a scientific setting, enabling automated yet interpretable generation of insights.

The key contributions of this work are: (1) a \textbf{scientific hybrid QA architecture} that couples executable spatiotemporal SQL programs over spectroscopy-derived peak parameters with literature retrieval and citation-grounded synthesis in a single interaction; (2) a \textbf{SUQL-style planner} that decomposes open-ended scientific questions into coordinated structured (SQL) and unstructured (vector search) operations, enabling multi-step phenomena such as cross-timestep comparisons and derived metrics (i.e., D/G) without requiring users to write queries; and (3) an \textbf{evaluation suite} for hybrid scientific assistants that combines SQL correctness, groundedness under varying retrieval depth, retrieval effectiveness (precision/recall/diversity), and expert ratings from domain scientists.

\section{Related Work}

\subsection{Raman Spectroscopy in Battery Research}
Raman spectroscopy has long been used to characterize battery materials. It can probe lattice structures, detect phase changes, and identify surface species via characteristic vibrational peaks \cite{baddour-hadjean_raman_2010}. For instance, the D (~1350 cm$^{-1}$) and G (~1580 cm$^{-1}$) bands of carbon report on relative disorder; the ratio of I(D)/I(G) > 1 (I = intensity) typically indicates significant disorder or defects in carbon \cite{ferrari_resonant_2001}. In battery electrodes, an increasing I(D)/I(G) ratio may signal carbon black degradation or binder decomposition. Meanwhile, metal-oxide cathodes exhibit Raman-active modes such as A1g and Eg; changes and shifts in the A1g peak intensity or position can reflect lithium de-intercalation or oxygen lattice distortion \cite{flores_situ_2018}. Operando Raman studies have observed phenomenon like spatially non-uniform lithiation, indicating heterogeneous battery material behavior and performance and hinting at underlying drivers of degradation modes \cite{hiraoka_advanced_2025}. To date, most analysis of operando Raman is manual or relies on static plotting; SpectraQuery is, to our knowledge, the first system to enable interactive querying of operando Raman data integrated with literature insight.

\begin{figure}[t]
\centering
  \includegraphics[width=0.9\columnwidth]{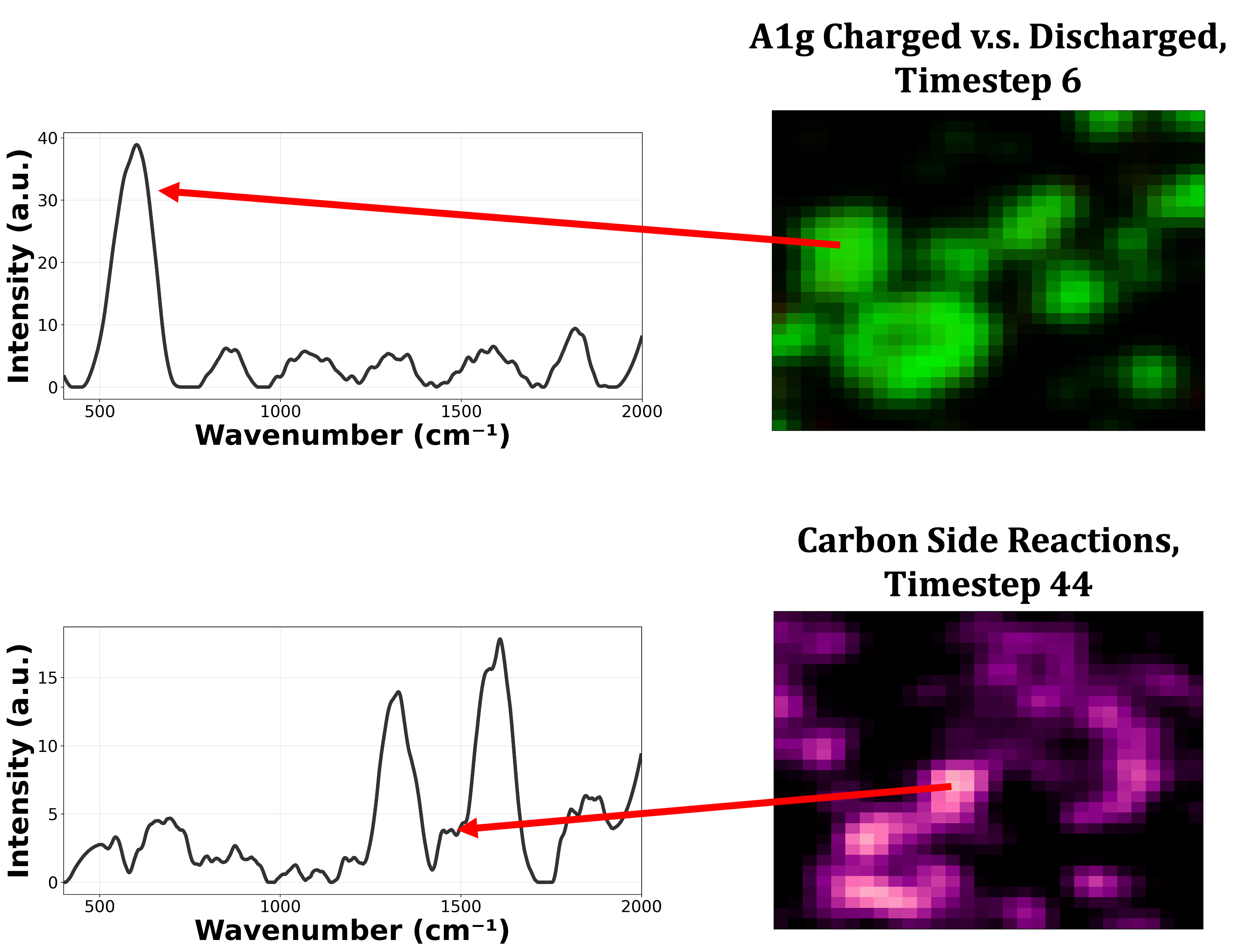}
  \caption{Each spatial pixel in the Raman maps (right panels) represents the intensity distribution of characteristic vibrational modes identified by automated peak detection. Representative spectra from individual pixels (left panels) show the corresponding local features—(top) A1g charged transition-metal–oxygen vibration and (bottom) combined unknown carbon peaks}
  \label{fig:operando}
\end{figure}

\subsection{Structured + Unstructured Data Querying}
Traditional text-to-SQL research has tackled converting natural language questions into SQL on relational databases \cite{yu_typesql_2018}. Recently, hybrid Question Answering (QA) datasets like HybridQA explicitly require reasoning over both tabular data and text \cite{chen_hybridqa_2020}. Models that combine a knowledge base with a text corpus have shown improved coverage on such QA tasks. In particular, the Structured and Unstructured Query Language (SUQL) pioneered at Stanford, which achieved impressive performance on the HybridQA dataset, has introduced a paradigm for LLM in-context semantic parsers that can handle hybrid data \cite{liu_suql_2023}. SpectraQuery follows this paradigm, treating the Raman spectra as a structured SQL database, and the battery literature as unstructured text.   

While SpectraQuery uses text-to-SQL internally, our target setting differs from classical text-to-SQL benchmarks: many queries require derived scientific quantities (i.e., peak ratios), spatiotemporal comparisons across timesteps, and mechanistic interpretation that is not present in the database. A single-shot SQL query may retrieve the correct numbers but cannot supply literature-grounded explanations, whereas literature QA alone cannot reliably compute joins, aggregations, and comparisons. SUQL provides an explicit planning abstraction to coordinate these operations and expose inspectable intermediate evidence.

\subsection{Retrieval-Augmented Generation (RAG)}
LLMs augmented with retrieval have reached state-of-the-art results in knowledge-intensive tasks \cite{lewis_retrieval-augmented_2020}. By fetching relevant text from an external body, RAG models avoid hallucinations by providing verified information and citing sources. Prior RAG frameworks have focused on open QA over extremely large external corpora, such as the ATLAS model with open QA over Wikipedia \cite{izacard_atlas_2023}. SpectraQuery instead builds a tailored literature index of domain-specific works; scientific QA systems such as SciQA and BioGPT have shown that incorporating this type of tailored literature corpus can meaningfully enrich explanations \cite{auer_sciqa_2023, luo_biogpt_2022}. 

\subsection{LLM Planning and Execution}
The idea of using an LLM as a controller to decide and execute actions has gained traction with approaches like ReAct and OpenAI's function-calling API, which allow the LLM to produce a programmatic plan as intermediate steps in reaching a final answer \cite{yao_react_2022}. SpectraQuery leverages this concept via a two stage process: in stage one, the SUQL planner (powered by an LLM prompt) generates a SQL query and a separate literature search query from the user's question. In step two, after retrieving numerical data and relevant textual passages, the LLM produces the final answer with the secured evidence. 

\section{Core Ideas and Methodology}

\subsection{System Overview}
SpectraQuery is a hybrid QA system that combines a relational database of Raman spectroscopy results with a vector-indexed corpus of literature. The user interacts via a chat-style interface, asking questions in natural language. Internally, the system comprises several components: a query planner (SUQL) that interprets the question, a SQL query executor, a semantic literature search, and an LLM answer synthesizer. The design follows a plan-execute-synthesize paradigm, enabling complex multi-hop reasoning in a single seamless interaction \cite{liu_suql_2023}. SpectraQuery's web interface, currently implemented with the Python toolkit Streamlit, allows the user to have a multi-turn conversation. A full receipt of all queries, intermediate steps and information retrieval, and final synthesized answers is also easily downloadable in a PDF, allowing for accessible sharing of results. 

\subsection{Structured Database}
\subsubsection{Raman Data Preprocessing}
The raw Raman dataset used for validation of SpectraQuery comes from line-scan operando experiments on a lithium-ion layered transition metal oxide (LTMO) cathode sampled at 114 timesteps during charge and discharge cycling (experiment performed at SLAC National Accelerator Laboratory). At each timestep, a linear spatial scan was performed, yielding a 30x30 two-dimensional grid of Raman spectra at each (x, y) coordinate (900 spectra per timestep). Each spectrum spans 100-2700 cm$^{-1}$, and contains many peaks of interest. 

Raw spectra underwent a standard preprocessing pipeline of spike removal, smoothing with a Savitzky-Golay filter (window of 31 points), and Asymmetric Least Squares baseline subtraction. The resulting smooth, baseline-corrected spectrum was then passed to a Bayesian peak fitting algorithm, yielding a set of detected peaks with fitted parameters. 

Each detected peak was assigned to one of eight canonical families by matching its fitted center to expected wavenumber ranges: TM–O lattice modes ($E_g\!\approx\!476$~cm$^{-1}$, $A_{1g,d}\!\approx\!534.5$~cm$^{-1}$, $A_{1g,c}\!\approx\!595.5$~cm$^{-1}$), carbon bands ($D\!\approx\!1330.5$~cm$^{-1}$ indicating disordered/sp$^3$-rich carbon and $G\!\approx\!1596.8$~cm$^{-1}$ indicating graphitic/sp$^2$ carbon), and three additional carbon features ($U_1\!\approx\!1173.3$~cm$^{-1}$, $U_2\!\approx\!1508.1$~cm$^{-1}$, $U_3\!\approx\!1564.0$~cm$^{-1}$); peaks with confidence $<0.98$ were discarded.

\begin{figure}[t]
\centering
  \includegraphics[width=\columnwidth]{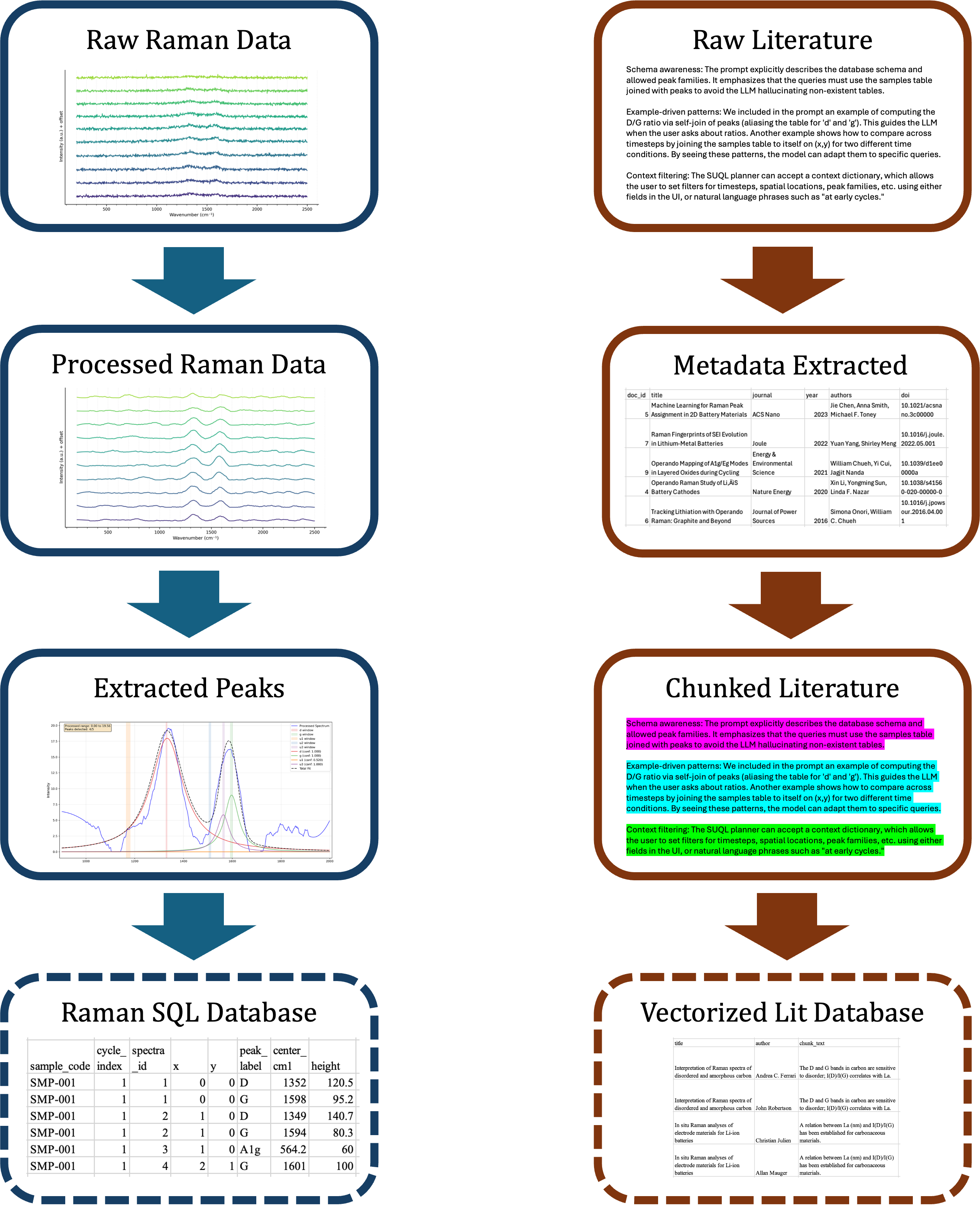}
\caption{Data and literature pipelines. (Left) Raman preprocessing \textrightarrow{} peak fitting \textrightarrow{} relational tables. 
  (Right) Literature ingestion \textrightarrow{}  metadata embeddings \textrightarrow{} chunking \textrightarrow{} vector search.}
  \label{fig:database}
\end{figure}

\subsubsection{Database Schema}
The processed data is stored in a SQLite relational database (via SQLAlchemy in our implementation). The process of building this Raman database is outlined on the left-hand side of Figure 2. The schema has two main tables: samples and peaks. 

samples(id, ts, x, y): Each row represents a measurement at a specific timestep (ts) and spatial position (x,y). We assign a unique ID to each (ts, x, y). 

peaks(id, sample\_id, family, center, height, width): Each detected peak is a row linked via sample\_id to a specific sample. The family field is a categorical string $\in$ \{a1g\_c, a1g\_d, eg, d, g, u1, u2, u3\}. If a peak family was not detected or filtered out at that sample, no row exists for that particular combination. 

\subsection{Unstructured Database}
In parallel to building the SQL database, we constructed an unstructured text corpus of battery literature. The process of building this literature database is outlined on the right-hand side of Figure 2. 50 PDF documents were selected, including review articles on Raman in batteries, research papers on cathodes, papers discussing battery degradation, foundational texts on Raman signatures, etc. These PDFs were extracted into text using PyMuPDF, and then split into overlapping chunks (~1000 tokens each with 150-token overlaps).

OpenAI's text-embedding-ada-002 model, through the LiteLLM embedding API, was then applied to generate embeddings for each chunk, and these resulting embeddings and chunks were stored in a ChromaDB persistent index. Metadata, such as source paper title, page number, and section header, were also stored for each chunk. The final vector store allows for fast similarity search: given a query embedding, it returns top-N chunks with cosine similarity scores.

\subsection{SUQL Query Planner}
\subsubsection{Core Functionality}
The core of SpectraQuery is the Structured-Unstructured Query Language (SUQL) planner. The process of this SUQL planner is outlined in Figure 3. The SUQL planner uses an LLM (this implementation used OpenAI GPT-4 via the LiteLLM API interface) to transform the user's natural language question into two outputs: 1) a SQL SELECT query with optional parameters for the structured data, and 2) a concise keyword query for the literature search. This essentially decomposes the question into "What data do we need?" and "What background do we need?" We crafted a system prompt for the planner with detailed instructions and provided a few-shot examples for complex query types. Key aspects of the SUQL planner design include:

\begin{figure}[t]
\centering
  \includegraphics[width=\columnwidth]{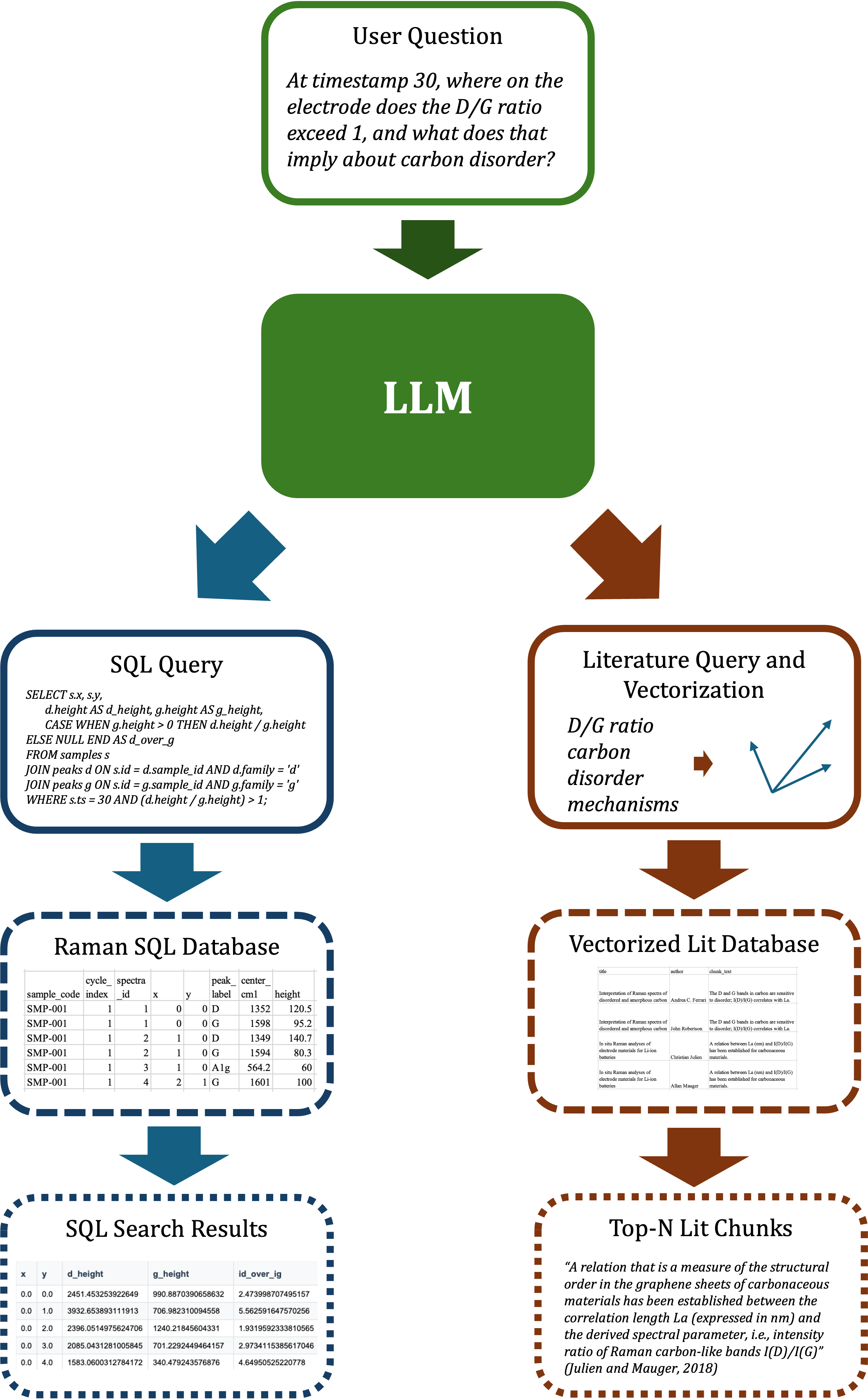}
\caption{The planner parses a natural-language question, emits coordinated SQL over the Raman database (left) and a literature query over the vector index (right), and returns harmonized intermediate tables/snippets for generation.}
  \label{fig:suql-path}
\end{figure}

Schema awareness: The prompt explicitly describes the database schema and allowed peak families. It emphasizes that the queries must use the samples table joined with peaks to avoid the LLM hallucinating non-existent tables. 

Example-driven patterns: We included in the prompt an example of computing the D/G ratio via self-join of peaks (aliasing the table for 'd' and 'g'). This guides the LLM when the user asks about ratios. Another example shows how to compare across timesteps by joining the samples table to itself on (x,y) for two different time conditions. By seeing these patterns, the model can adapt them to specific queries. 

Context filtering: The SUQL planner can accept a context dictionary, which allows the user to set filters for timesteps, spatial locations, peak families, etc. using either fields in the UI, or natural language phrases such as "at early cycles."

Safety checks: After the LLM drafts the SQL, a validation function is run to ensure it is a safe SELECT-only query and only references the two allowed tables. Disallowed keywords such as DROP or DELETE trigger a rejection and error.

\subsubsection{Parallel Query Execution}
Once the SUQL planner returns the SQL and literature query, SpectraQuery executes them in parallel. The SQL query is run on the SQLite database, resulting in a set of rows which are formatted and possibly truncated for easier viewing. The literature query is vectorized using the aforementioned OpenAI text-embedding-ada-002 model, and a similarity search is performed in ChromaDB. The top k passages are retrieved, with text snippets and source metadata. Each passage comes with its text snippet and source metadata.

\subsection{Answer Synthesis}
The final step is to synthesize an answer that combines the structured data results and literature context. The process of this synthesis is outlined in Figure 4. We use the LLM in a second stage prompt for this. The prompt template instructs: summarize the relevant findings from the data, incorporate relevant points from the literature snippets, and produce a coherent answer. To maintain transparency and user trust, we also ask it to cite sources: for data, we cite it as "(Data: ...)" and for literature we cite by title.

\begin{figure}[t]
  \includegraphics[width=\columnwidth]{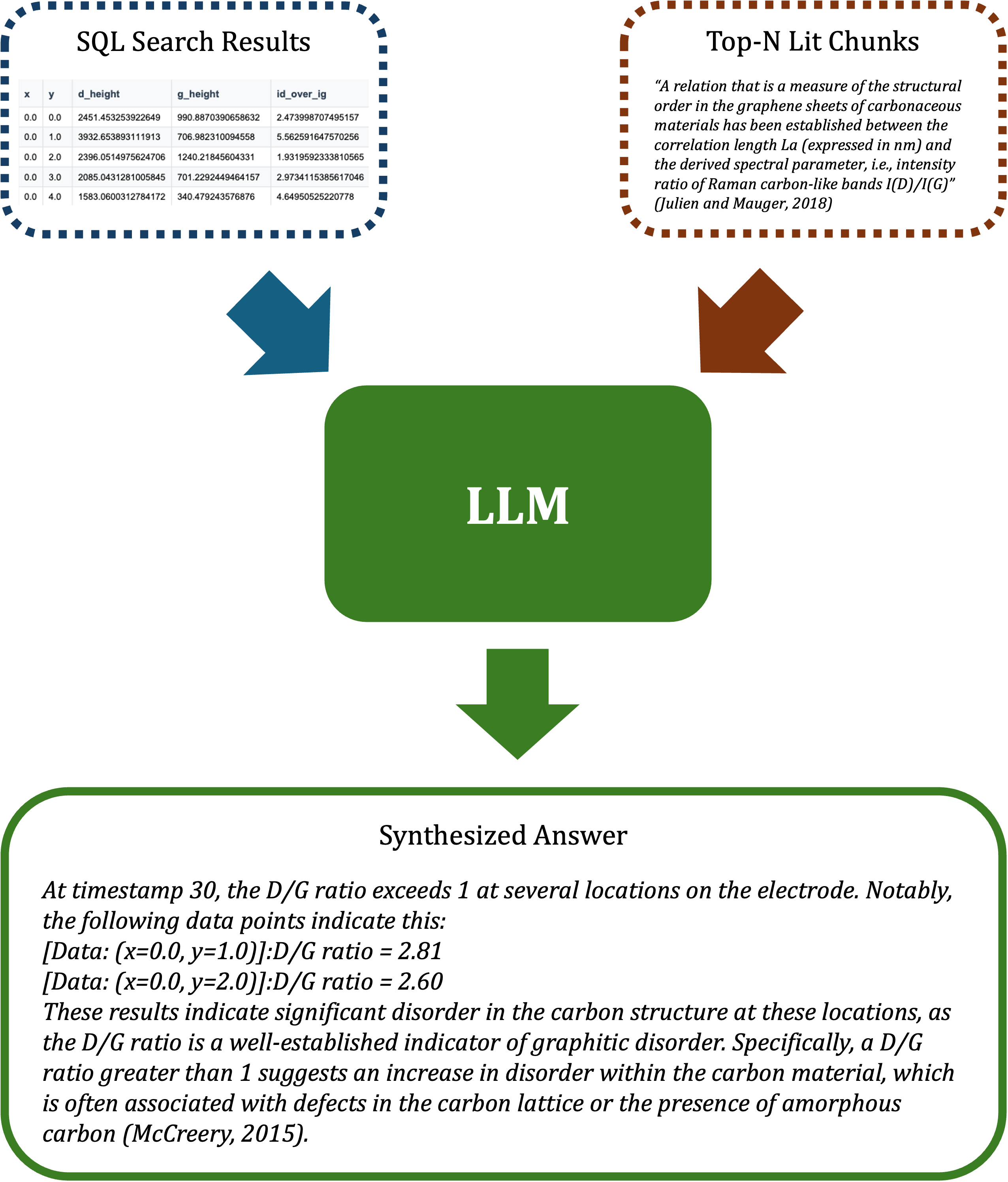}
\caption{Structured outputs (i.e., D/G table; left) and literature snippets (right) are fused by the LLM to produce a grounded, cited natural-language answer (bottom).}
  \label{fig:answer}
\end{figure}

\section{Experimental Results}

We evaluate SpectraQuery against three baseline configurations that isolate
key components of the system: RAG-only (literature retrieval + synthesis, no SQL),
SQL-only (execute generated SQL, no literature or synthesis), and text-to-SQL
(a single SQL query generated directly from the question). These baselines reflect existing approaches commonly used in scientific
question answering and data analysis. We assess SpectraQuery along four axes:
(1) the correctness of SUQL-generated SQL queries,
(2) the factual groundedness of synthesized answers,
(3) literature retrieval effectiveness, and
(4) expert perceptions of the final responses.
Unless otherwise noted, all evaluations are performed on a benchmark set of
30 expert-curated Raman–battery questions derived from the operando LTMO
cathode dataset (full list in Appendix~\ref{sec:benchmark-questions}).

\subsection{LLM-As-A-Judge Evaluation}

\subsubsection{SQL Query Accuracy}
\begin{figure}[t]
  \centering
  \includegraphics[width=\linewidth]{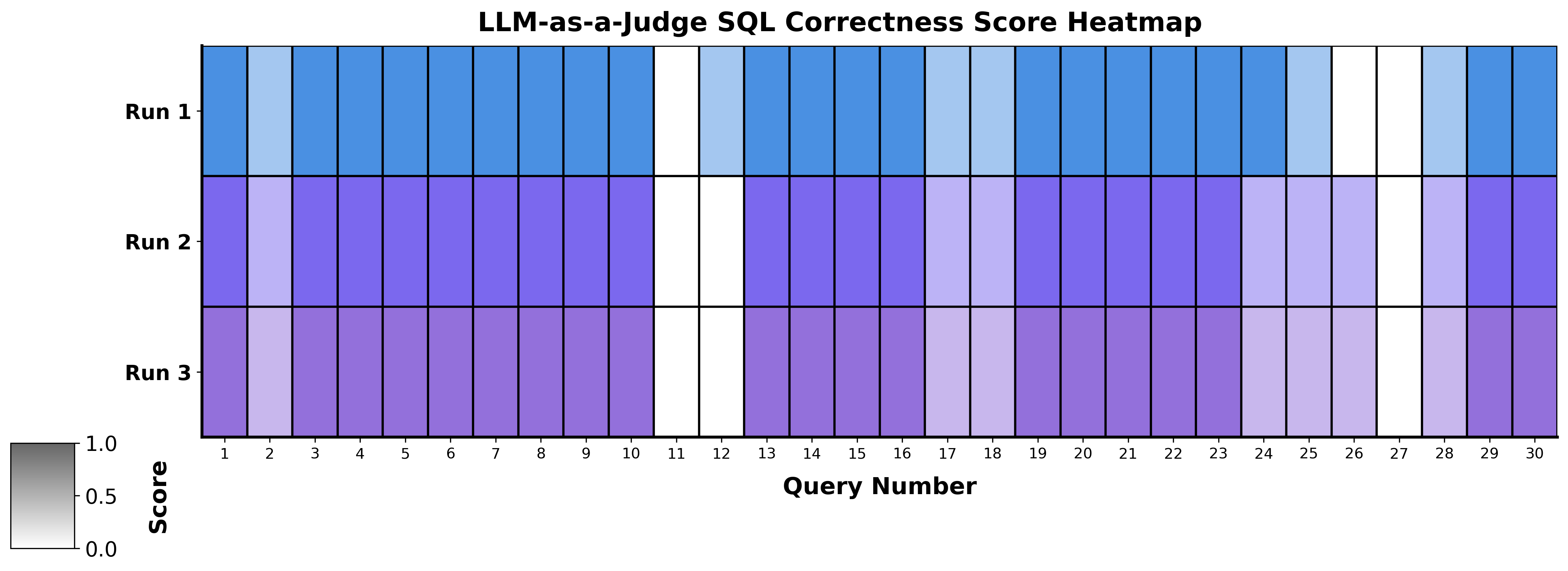}
  \caption{LLM-as-a-judge SQL correctness scores (0, 0.5, 1.0) for three independent runs across the 30 benchmark questions. Darker cells indicate higher correctness.}
  \label{fig:sql-heatmap}
\end{figure}

We evaluated the correctness of SUQL-generated SQL using an LLM-as-a-judge protocol \cite{zheng2023judging}. For each benchmark question, we logged the natural-language prompt, the generated SQL, execution status, and returned rows, and asked GPT-5 to score query correctness on a three-level rubric: \textbf{1.0} if the SQL executes and fully satisfies the question (correct tables/joins/filters/aggregations; results match the requested information), \textbf{0.5} if it executes but only partially satisfies the question (i.e., missing a required condition or returning an incomplete slice), and \textbf{0.0} if it fails to execute or fundamentally misinterprets the question. To assess robustness, we repeated scoring three times with independent GPT-5 calls; across 30 questions, the fraction of queries scored fully correct (1.0) was \textbf{80.0\%} (Run 1), \textbf{78.3\%} (Run 2), and \textbf{78.3\%} (Run 3). The remaining errors were split between partial and incorrect cases, typically due to missing one of several requested conditions (i.e., checking A$_{1g}$ change but not D/G), flipped inequality directions, or incorrect joins between \texttt{samples} and \texttt{peaks} for cross-timestep comparisons. Figure~\ref{fig:sql-heatmap} shows that these failures are concentrated in a small subset of more complex questions.

\subsubsection{Synthesized Answer Groundedness}
\begin{figure}[t]
  \centering
  \includegraphics[width=\linewidth]{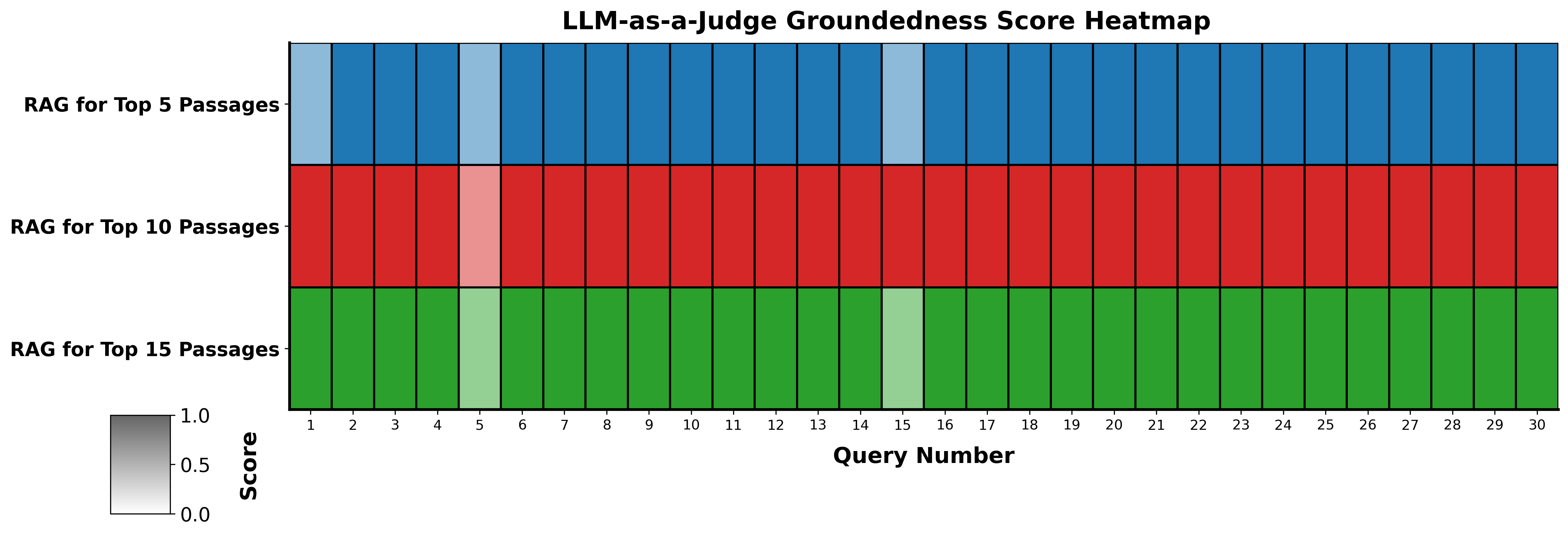}
  \caption{LLM-as-a-judge groundedness scores for synthesized answers when conditioning on the top-5, top-10, and top-15 retrieved passages.}
  \label{fig:groundedness-heatmap}
\end{figure}

We next measured how well final answers are grounded in retrieved evidence \cite{zheng2023judging}. For each question, we concatenated the SQL result table with the top-$k$ retrieved literature passages and asked GPT-5 to assign a groundedness score on a discrete rubric: \textbf{1.0} if all claims are supported by the provided context, \textbf{0.5} if the answer is partially supported but includes unsupported or speculative statements, and \textbf{0.0} if key claims cannot be justified from the context. Varying the number of retrieved passages shows a clear retrieval--grounding tradeoff: \textbf{83.3\%} of answers are fully grounded with top-5 passages, rising to \textbf{93.3\%} with top-10 and \textbf{96.7\%} with top-15. Gains from 10 to 15 passages are smaller, suggesting diminishing returns beyond roughly 10 passages. Residual non-grounded cases are primarily due to missing retrieval (relevant mechanisms absent from the retrieved set) or mild over-generalization beyond the evidence. Figure~\ref{fig:groundedness-heatmap} summarizes per-query groundedness across $k$.

\subsection{Retrieval Effectiveness Metrics}

To better understand the retrieval component itself, we measured standard information retrieval metrics against a “gold” set of relevant papers \cite{singhal2001modern}. For each of the 30 benchmark questions, we formed an expert-curated ranked list of the top five relevant ground-truth papers from our corpus that best addressed the phenomenon of interest. We then ran SpectraQuery’s literature retriever and computed paper-level \textbf{Precision@k} (the fraction of retrieved papers in the top $k$ that appear in the ground-truth set) and \textbf{Recall@k} (the fraction of ground-truth papers that appear within the top $k$ retrieved papers) \cite{manning2008introduction}, along with \textbf{UniqueDocs@k} (the number of distinct source papers represented among the top $k$ retrieved passages, as a proxy for diversity) \cite{clarke2008novelty}.

\begin{table}[t]
  \centering
  \begin{tabular}{lccc}
    \hline
    \textbf{Metric} & \textbf{@1} & \textbf{@3} & \textbf{@5} \\
    \hline
    Precision   & 0.567 & 0.556 & 0.580 \\
    Recall      & 0.433 & 0.533 & 0.600 \\
    Unique Docs &   --  & 1.80  & 2.30 \\
    \hline
  \end{tabular}
  \caption{Retrieval performance aggregated over 30 benchmark queries.}
  \label{tab:retrieval-metrics}
\end{table}

\begin{figure}[t]
  \centering
  \includegraphics[width=\linewidth]{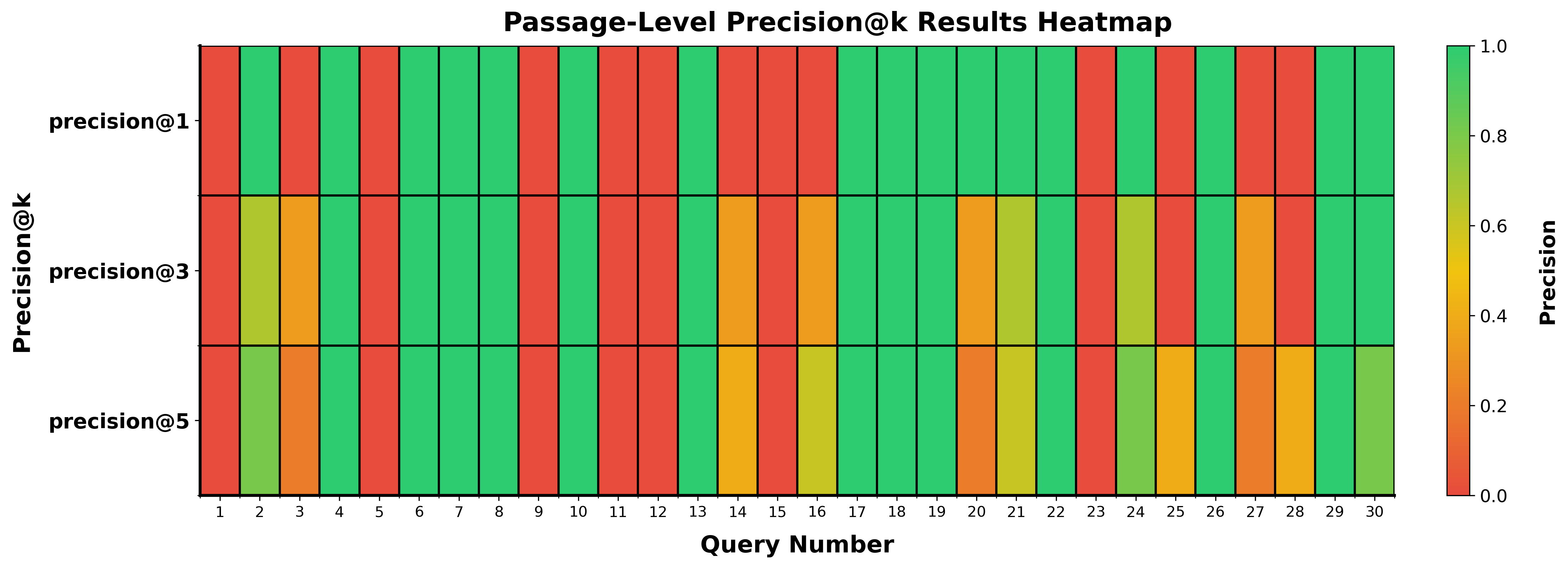}

  \vspace{0.75em}

  \includegraphics[width=\linewidth]{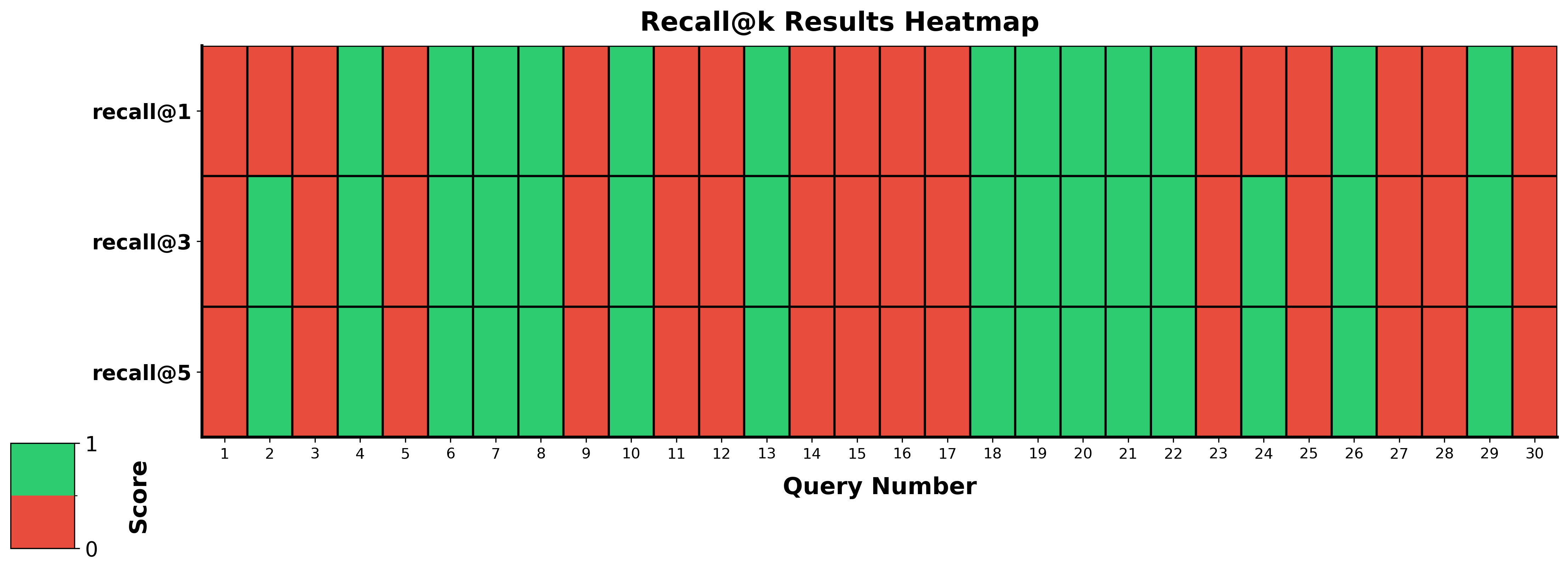}

  \vspace{0.75em}

  \includegraphics[width=\linewidth]{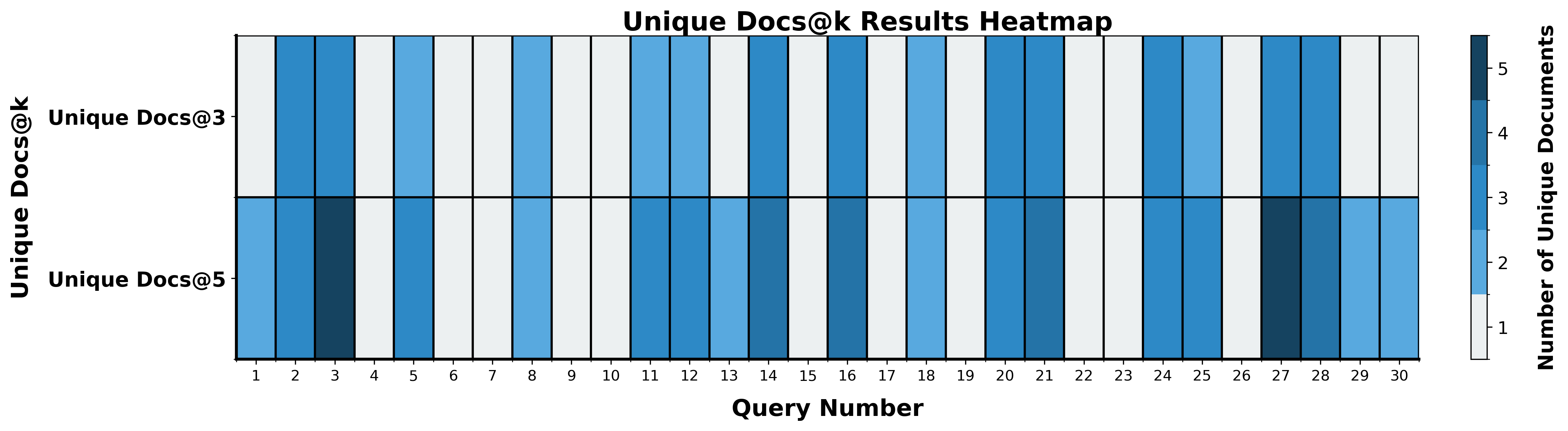}

  \caption{Per-query retrieval effectiveness visualized via Precision@k (top), 
  Recall@k (middle), and UniqueDocs@k (bottom) across the 30 
  benchmark questions. Together, these heatmaps show that SpectraQuery retrieves 
  highly relevant but somewhat redundant papers, with recall failures concentrated 
  on a small subset of harder questions.}
  \label{fig:retrieval-heatmaps}
\end{figure}

Averaged across queries, we obtain the results in Table~\ref{tab:retrieval-metrics}. These values indicate that when the retriever retrieves a paper, it is frequently relevant (precision around 0.56–0.58). Recall is more modest: by $k=5$, we recover on average 60\% of the manually identified relevant papers. UniqueDocs@k reveals that the top-$k$ passages often cluster on a small subset of documents (i.e., only 1.8 distinct papers among the top 3 passages).

Figure~\ref{fig:retrieval-heatmaps} provides a more fine-grained view across all 30 questions. Panel (a) visualizes passage-level Precision@k, which largely mirrors the paper-level trends and confirms that most retrieved passages are either clearly relevant (score 1.0) or clearly irrelevant (0.0). Panel (b) shows that recall failures are concentrated on a handful of queries whose information need is broad or abstract. Panel (c) highlights that many queries have UniqueDocs@3 = 1 and UniqueDocs@5 close to 2, indicating limited diversity: the retriever tends to return multiple passages from a single favored paper instead of surfacing complementary sources. This behavior explains some of the groundedness errors, since missing diversity can deprive the LLM of alternative mechanistic explanations.

\subsection{Expert Feedback Evaluation}

\begin{figure}[t]
  \centering
  \includegraphics[width=\linewidth]{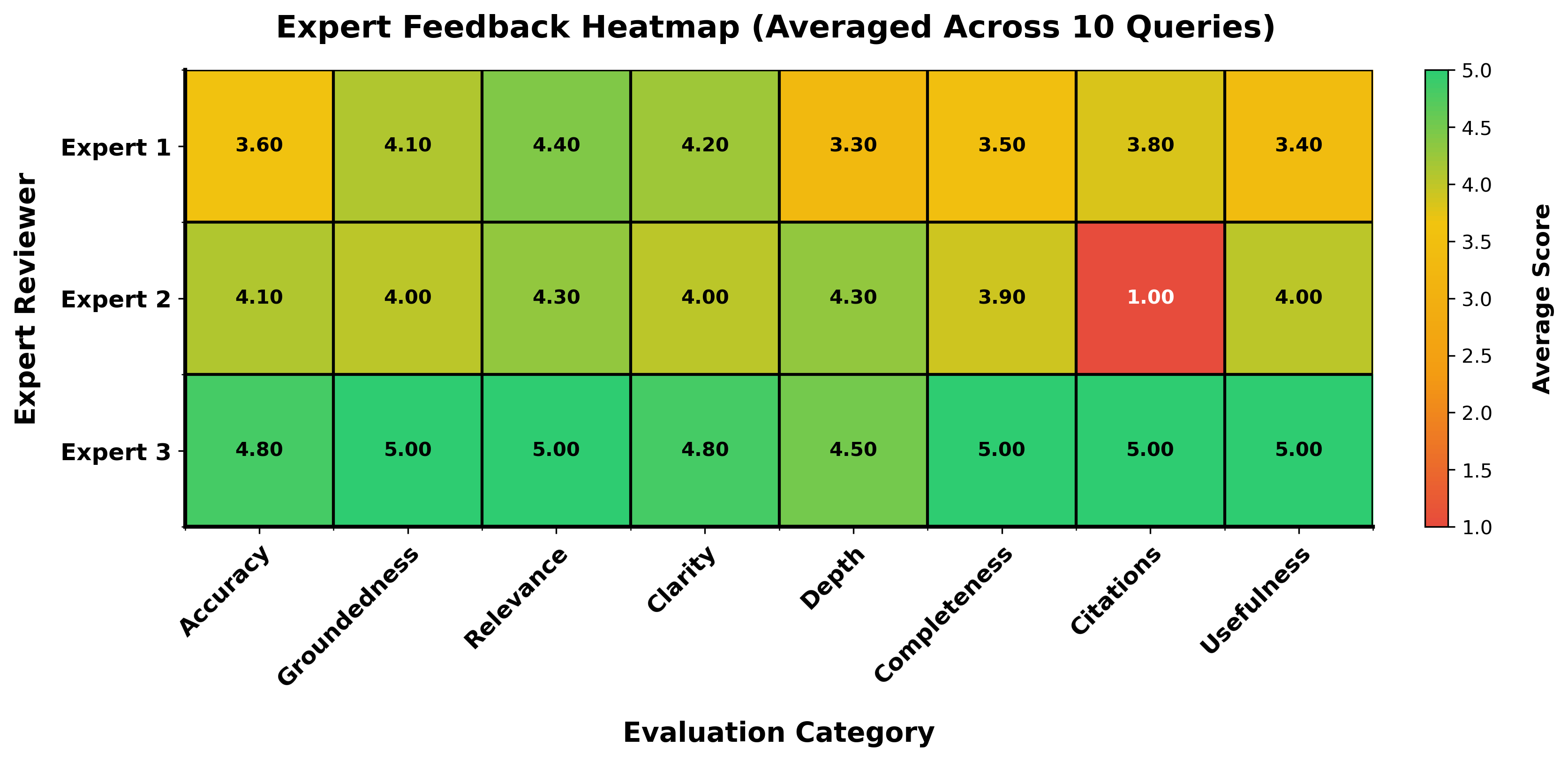}
  \caption{Expert feedback heatmap (10 questions averaged per cell). Each row corresponds to one reviewer and each column to one evaluation category, with numerical Likert scores overlaid.}
  \label{fig:expert-heatmap}
\end{figure}

To complement automated metrics, we collected expert feedback from three battery scientists at the SLAC National Accelerator Laboratory \cite{van2024field, semnani2025detecting}. Ten representative SpectraQuery answers were rated on a 1--5 Likert scale across eight dimensions, including scientific accuracy, grounding, relevance, clarity, depth, completeness, citation interpretability, and overall usefulness. Two exact query–answer pairs shown to experts are reproduced in Appendix~\ref{sec:expert-packet}. Table~\ref{tab:expert-ratings} reports mean scores over 30 judgments (10 answers $\times$ 3 reviewers), while Figure~\ref{fig:expert-heatmap} visualizes per-reviewer scoring patterns.

Overall, experts rated the system favorably. Core quality metrics—accuracy, grounding, relevance, and clarity—are all above 4.0, indicating that responses are generally correct, well-supported, and directly address the questions. Depth and completeness scores near 4.0 suggest that the system typically captures the key mechanisms experts expect. The lowest-scoring dimension is citation interpretability (3.27), reflecting difficulty tracing references to specific sources or sections, and some variability across reviewers. Taken together, expert feedback validates SpectraQuery’s scientific credibility while highlighting retrieval completeness and citation presentation as primary areas for improvement.

\begin{table}[t]
  \centering
  \begin{tabular}{l c}
    \hline
    \textbf{Metric} & \textbf{Mean Score (1--5)} \\
    \hline
    Scientific Accuracy & 4.17 \\
    Grounding in Evidence & 4.37 \\
    Relevance & 4.57 \\
    Clarity & 4.33 \\
    Depth of Insight & 4.03 \\
    Completeness & 4.13 \\
    Interpretability of Citations & 3.27 \\
    Overall Usefulness & 4.13 \\
    \hline
  \end{tabular}
  \caption{Average expert ratings for 10 SpectraQuery answers. Each score is averaged across 3 reviewers (30 ratings total).}
  \label{tab:expert-ratings}
\end{table}

\section{Insights and Discussion}
Across automated and expert evaluations, SpectraQuery is most reliable when the retriever surfaces the right mechanisms and diverse supporting sources. When key evidence is missing, answers become incomplete or over-generalize despite correct SQL execution. This coupling is consistent with our retrieval metrics (Recall@5 = 0.60; low diversity with $\sim$1--2 unique documents in the top-$k$), which often return multiple passages from a single favored paper. Improving recall and diversity via reranking or multi-stage retrieval is therefore the most direct lever for improving scientific reliability.

Furthermore, we observed that LLM-as-a-judge provides a scalable diagnostic for SQL correctness and groundedness, but experts apply stricter standards for completeness and citation usability. In practice, automatic judging is effective for ablations and error-finding, while expert review remains necessary to assess whether responses meet domain expectations for interpretability and scientific nuance.

\section{Conclusions and Future Work}

We presented SpectraQuery, a hybrid retrieval-augmented conversational system that allows researchers to pose natural-language questions over operando Raman datasets and scientific literature simultaneously. By combining a SUQL-inspired planner, a structured SQL backend over peak-parameter databases, and a domain-specific literature retriever, SpectraQuery produces cited answers that unify numerical evidence with mechanistic explanations. To our knowledge, this is one of the first end-to-end systems that treats operando spectroscopy data and battery papers as a single queryable space.

Our evaluation shows that the approach is both feasible and useful. LLM-as-a-judge experiments indicate that roughly 80\% of generated SQL queries are fully correct, and groundedness scores exceed 90\% when the model is conditioned on sufficiently many retrieved passages. Expert battery scientists rate SpectraQuery’s answers highly on accuracy, grounding, relevance, and clarity, with somewhat lower scores on citation interpretability and completeness. These results suggest that hybrid RAG architectures can meaningfully support real scientific workflows in niche domains, like battery spectroscopy. At the same time, our experiments highlight clear limitations; several avenues for improvement and extension are available.

First, SpectraQuery can be broadened beyond Raman spectra and text to incorporate additional battery-relevant modalities such as electrochemical time-series (voltage profiles, impedance), optical microscopy images, and other characterization datasets (i.e., X-ray diffraction), enabling queries like “Compare Raman observations with capacity loss” or “What side reactions are occurring in dark cathode regions,” with the planner orchestrating retrieval across structured metrics, image-analysis outputs, and literature. Second, the system could expand from a local literature index to external knowledge bases—for example, integrating open resources like the Materials Project to retrieve reference spectra and expected peaks (i.e., lithium carbonate) directly from aggregated community data \cite{horton_accelerated_2025}. Third, the LLM stack can be improved for cost and reliability: while GPT-5 is strong, smaller open models such as LLaMA-2 could be fine-tuned on domain-specific QA pairs to improve accuracy and reduce hallucinations \cite{touvron_llama_2023}, and complemented with a lightweight verification pass that checks each generated claim against retrieved evidence. Fourth, SpectraQuery could expose more user-facing tooling, including an optional interactive “debug mode” that allows expert users to inspect and edit intermediate SQL and retrieval results before synthesis, and richer visualization capabilities such as spectra overlays across timesteps, coordinate-linked plots, and spatial heatmaps (i.e., “Show me the D/G ratio map at t=30”) to make outputs more interpretable and actionable. Finally, our current implementation executes SQL and literature retrieval in parallel. A promising extension is an \textit{iterative two-pass} strategy: preliminary SQL findings (i.e., unexpected spatial hotspots or peak shifts) can trigger a refined literature query, and retrieved mechanisms can in turn suggest additional structured features to compute. This closed-loop interaction could reduce missed mechanisms and improve completeness on hard queries.

In summary, this work contributes an evaluation methodology that combines automated LLM-as-a-judge scoring with targeted expert review. We find that strong LLM judges correlate well with experts on obvious factuality errors but miss finer-grained concerns about completeness and presentation. Future AI assistants for scientific domains will likely need both scalable automatic metrics and carefully designed human studies to ensure that they truly augment, rather than mislead, domain experts. SpectraQuery is a first step in this direction, illustrating how structured and unstructured data can be jointly leveraged to accelerate insight generation in battery research.

\section{Acknowledgments}
This work was supported by the Knight-Hennessy Scholars Program, the Quad Fellowship, and the EDGE Fellowship. Research and experimental validation were conducted at the SLAC National Accelerator Laboratory Battery Center, which provided access to laboratory infrastructure and instrumentation. The authors also acknowledge support from the Assistant Secretary for Energy Efficiency and Renewable Energy (EERE), Office of Vehicle Technologies (VTO) of the US Department of Energy (DOE) under the Battery Materials Research (BMR) Program and Battery 500 Consortium.

The authors gratefully acknowledge the William Chueh Group, specifically Donggun Eum, and the SLAC National Laboratory Battery Center for their operando line-scan Raman data collection used in this study. We also thank Professor Monica Lam and Arjun Jain for their insightful discussions and for providing additional compute resources and feedback on the system design.

Some of the computing for this project was performed on the Stanford Sherlock cluster. We thank Stanford University and the Stanford Research Computing Center for providing computational resources and support that contributed to these research results. 

\section*{Limitations}

SpectraQuery has several limitations that motivate future work. 
First, the system’s reliability depends on retrieval quality: while precision is relatively high, recall and document diversity are imperfect, meaning that relevant mechanisms may be absent from the retrieved context for complex or niche queries. In such cases, answers may be incomplete or rely too heavily on a single source.

Second, structured reasoning depends on the correctness of SUQL-generated SQL. Although most queries are correct, errors such as missing conditions or incorrect joins can distort numerical summaries if undetected. The current system mitigates this risk through evaluation and filtering, but does not yet perform full semantic verification of query intent.

Third, SpectraQuery is evaluated on a single operando Raman dataset and a curated literature corpus. While the architecture is general, performance may vary across materials systems, experimental modalities, or scientific domains with different data distributions or terminology.

Fourth, portions of the programming and software development were performed with the assistance of AI-based tools, including GPT-4 and the Cursor development environment, which were used as productivity aids during code implementation and debugging. GPT-4 was also used to support editing and refinement of the manuscript text. All experimental design choices, evaluations, and interpretations were authored and verified by the authors, but the use of such tools may introduce subtle biases that are difficult to fully quantify.

Fifth, we do not provide a full quantitative head-to-head comparison against proprietary long-context assistants or end-to-end text-to-SQL systems, because many baselines do not expose comparable intermediate artifacts (executed SQL, retrieved passages) needed for our correctness and grounding evaluations. Instead, we provide conceptual baselines aligned with the benchmark requirements and emphasize reproducible component-level metrics.

Finally, expert evaluation involved three domain experts who are professional colleagues of the authors. All reviewers provided explicit consent for participation and for the use of their anonymized feedback in this study. The limited number of reviewers and their professional proximity to the authors may constrain the generalizability of the qualitative findings.

\section*{Ethical Considerations and Potential Risks}

SpectraQuery is designed as a research assistant to support scientific interpretation of experimental battery data and literature, rather than as an autonomous decision-making system. As such, its primary risks arise from misinterpretation or over-reliance on generated explanations. If used without expert oversight, incomplete retrieval or incorrect SQL could lead to misleading conclusions about material behavior or degradation mechanisms.

The system does not generate new experimental data, access personal information, or operate on sensitive human data. All inputs consist of laboratory measurements and publicly available scientific literature. Nevertheless, users should treat generated explanations as hypotheses supported by retrieved evidence, not as definitive conclusions.

A secondary risk concerns overgeneralization. The system is evaluated on a specific materials system and literature corpus, and its outputs may not transfer reliably to other chemistries or experimental settings. We mitigate these risks by emphasizing interpretability, explicit citation of sources, and positioning SpectraQuery as a decision-support tool rather than a substitute for domain expertise.

Finally, while AI-based tools were used to assist code development and writing, all scientific claims and evaluations were reviewed and validated by the author. The system is intended for research use only, and not for deployment in safety-critical or operational battery management settings.

\bibliography{custom}

\appendix

\clearpage
\section{Benchmark Question Set}
\label{sec:benchmark-questions}

To evaluate SpectraQuery's SQL correctness, retrieval quality, groundedness, 
and expert-rated usefulness, we curated 30 Raman–battery analysis questions 
representing real scientific workflows of existing collaborators.  
Questions probe peak evolution, degradation signatures, spatial heterogeneity, 
unknown-mode behavior, and mechanistic interpretation.

The benchmark set is divided below into thematic categories. 

\subsection{A. Expert Evaluation Subset (Questions 1–10)}
These ten questions span core Raman diagnostics used by domain scientists 
and were selected for blinded expert scoring of accuracy, clarity, grounding, 
and usefulness.

\begin{enumerate}
    \item Which timestep has the highest average A1g charged height, and what does that imply about the state of charge at that timestep?
    \item Which timestep has the highest D/G ratio, and what does that imply about carbon disorder?
    \item What is the single highest A1g discharged peak, and what does that mean?
    \item Find coordinates/timestamps where u2 height $>$ 200 and discuss literature connecting such carbon-region unknowns to side products and capacity fade.
    \item What is the ratio of the average A1g charged height to the average A1g discharged height at timestep 60? What does this ratio mean?
    \item Which $(x,y)$ coordinate has the highest average D height, and what does that mean for carbon disorder?
    \item Which timestep has the single coordinate with the largest sum of u1, u2, and u3 peaks, and what does that indicate about side reactions?
    \item What is the difference between the average u3 height at the last timestep and at the first timestep, and what does this tell us about degradation due to cycling?
    \item At timestep 30, what is the A1g$_c$ height at location (0,0)? What is the A1g$_c$ height at location (15,15)? What does this difference mean about behavior on the cathode edge versus in the middle?
    \item Find timesteps and coordinates where the u3 height $>$ 200, and discuss what this means with respect to side reactions and lithium loss.
\end{enumerate}

\subsection{B. Peak Shape, Lattice Disorder, and Structural Transition Questions}

\begin{enumerate}\setcounter{enumi}{10}
    \item Compare the A1g$_c$ FWHM at the first and last timesteps. What does the broadening tell us about lattice disorder accumulation?
    \item Compute the A1g$_c$/E$_g$ intensity ratio at each timestep. At which timestep is the ratio lowest, and what structural transition is associated with this?
    \item Find any coordinates where the A1g$_c$ center $>$ 595 cm$^{-1}$. Summarize literature describing A1g blue-shifts during high-voltage oxygen activity.
    \item Determine where the A1g$_c$ and A1g$_d$ peaks disappear entirely. What failure mechanism does complete A1g loss correspond to?
    \item At timestep 75, which 10 coordinates have the highest A1g$_c$ intensity, and what might this say about spatially selective retention of TM–O structural integrity?
    \item At which timestep does the average G-band height reach a maximum? What does this mean about graphitic ordering?
\end{enumerate}

\subsection{C. Spatial Heterogeneity and Electrode-Asymmetry Questions}

\begin{enumerate}\setcounter{enumi}{16}
    \item Compare the D/G ratios for the left vs.\ right halves of the electrode at timestep 40. What spatial asymmetry does this reveal?
    \item Which timestep has the lowest D/G ratio, and what does that indicate about carbon structural recovery (if any)?
    \item At coordinates where D $>$ 400, retrieve literature linking high-D features to electrolyte attack or carbon amorphization.
    \item Identify any coordinates where the G-band disappears but A1g persists. What edge-case failure or delamination does this pattern map onto?
    \item Determine all points where u2 and u3 are simultaneously $>$150. What multipeak signatures are associated with multi-step side reactions?
    \item Find coordinates where u3 height $>$ A1g$_c$ height. What does dominance of unknown/byproduct modes imply about end-of-life chemistry?
    \item Compare A1g intensity at coordinates (0,0), (15,15), and (29,29). Which region ages fastest?
    \item Determine which quadrant of the electrode has the highest mean u2 height at the last timestep. What could be causing quadrant-specific side reactions?
    \item Find edge vs.\ center contrast in A1g FWHM at the last timestep. Why does the literature say edges degrade faster?
\end{enumerate}

\subsection{D. Degradation Kinetics and Mechanistic Interpretation Questions}

\begin{enumerate}\setcounter{enumi}{25}
    \item Compute the percentage loss of A1g intensity from timestep 0 $\rightarrow$ final. Interpret this loss in the context of typical TM–O bond destabilization percentages.
    \item Determine when the A1g discharged peak first falls below 50\% of its initial value, and explain the mechanistic significance.
    \item Compute the A1g$_\text{charged}$/A1g$_\text{discharged}$ ratio across all timesteps and find when it deviates most from unity. How does this ratio map to irreversibility?
    \item For all timesteps where u2 $>$ 150, retrieve literature that links mid-frequency Raman features in LTMO cathodes to parasitic oxygen reactions.
    \item Return all timesteps where average A1g$_c$ $<$ average D. What scenario does ``carbon dominating over lattice'' represent?
\end{enumerate}

\clearpage
\section{Expert Evaluation Packet}
\label{sec:expert-packet}

The following pages contain the first two SpectraQuery query–answer
examples provided to expert evaluators. These were
used to assess scientific accuracy, grounding, clarity, completeness, and
overall usefulness of the system.

\clearpage

\begin{figure}[t]
  \centering
  \includegraphics[width=\linewidth]{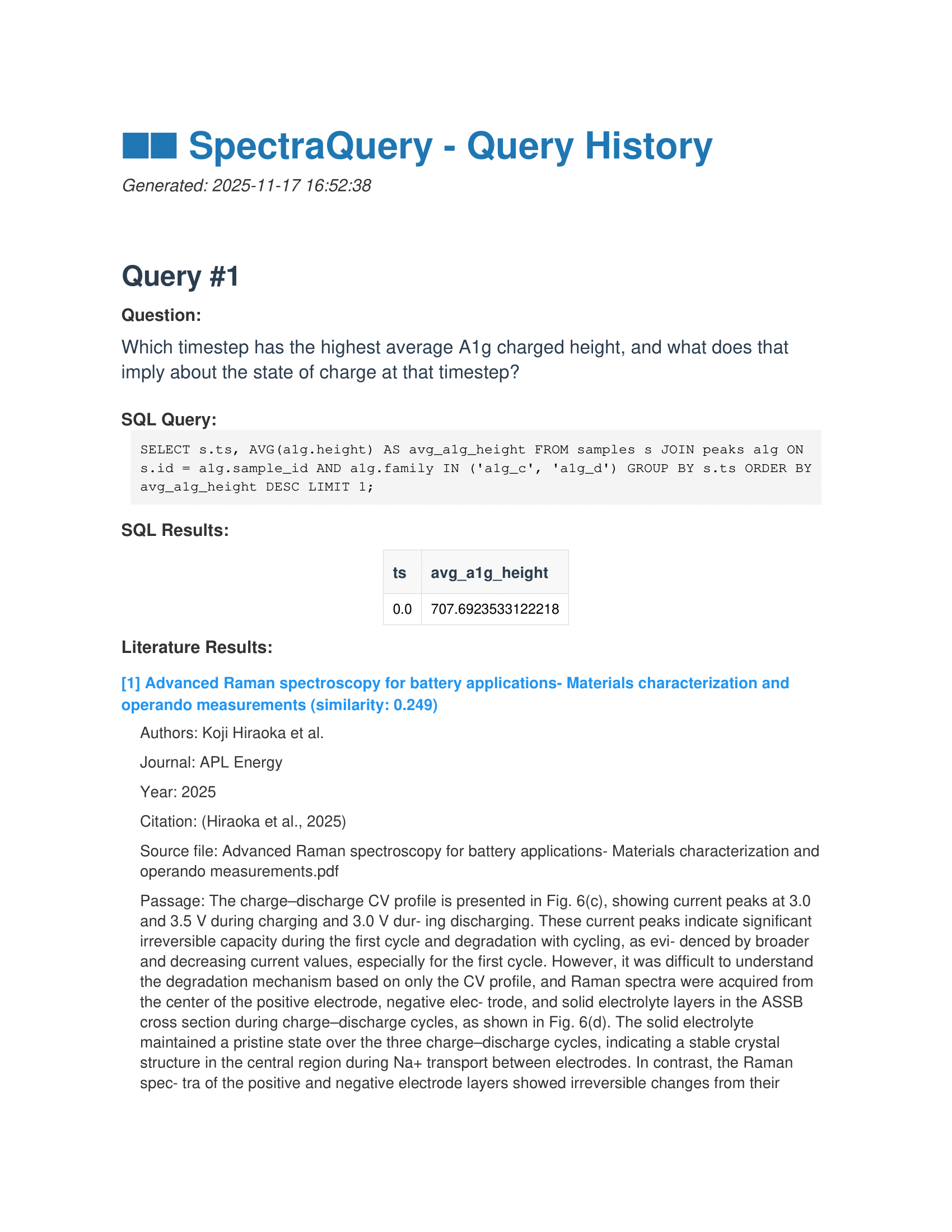}
  \caption{Expert evaluation packet, page 1.}
\end{figure}

\begin{figure}[t]
  \centering
  \includegraphics[width=\linewidth]{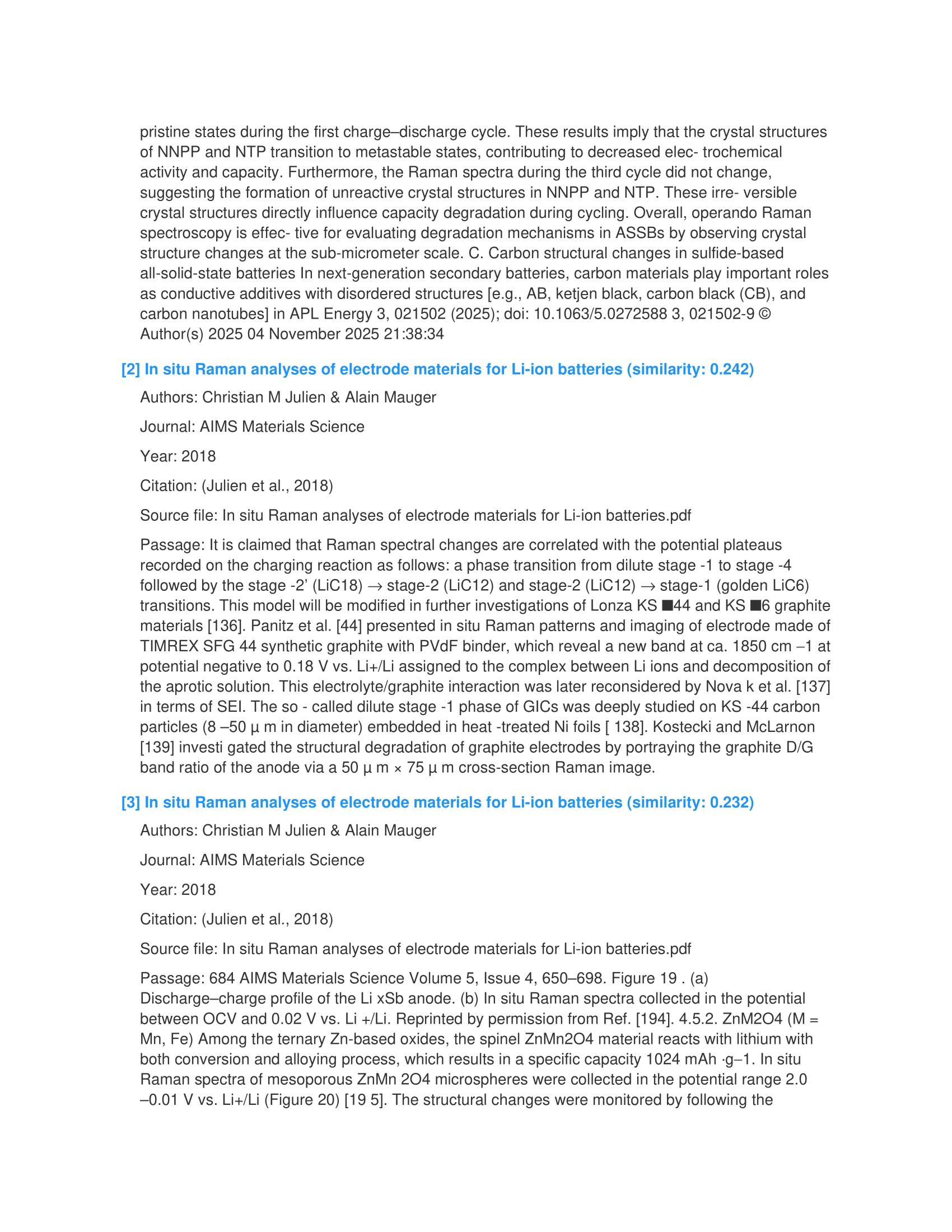}
  \caption{Expert evaluation packet, page 2.}
\end{figure}

\begin{figure}[t]
  \centering
  \includegraphics[width=\linewidth]{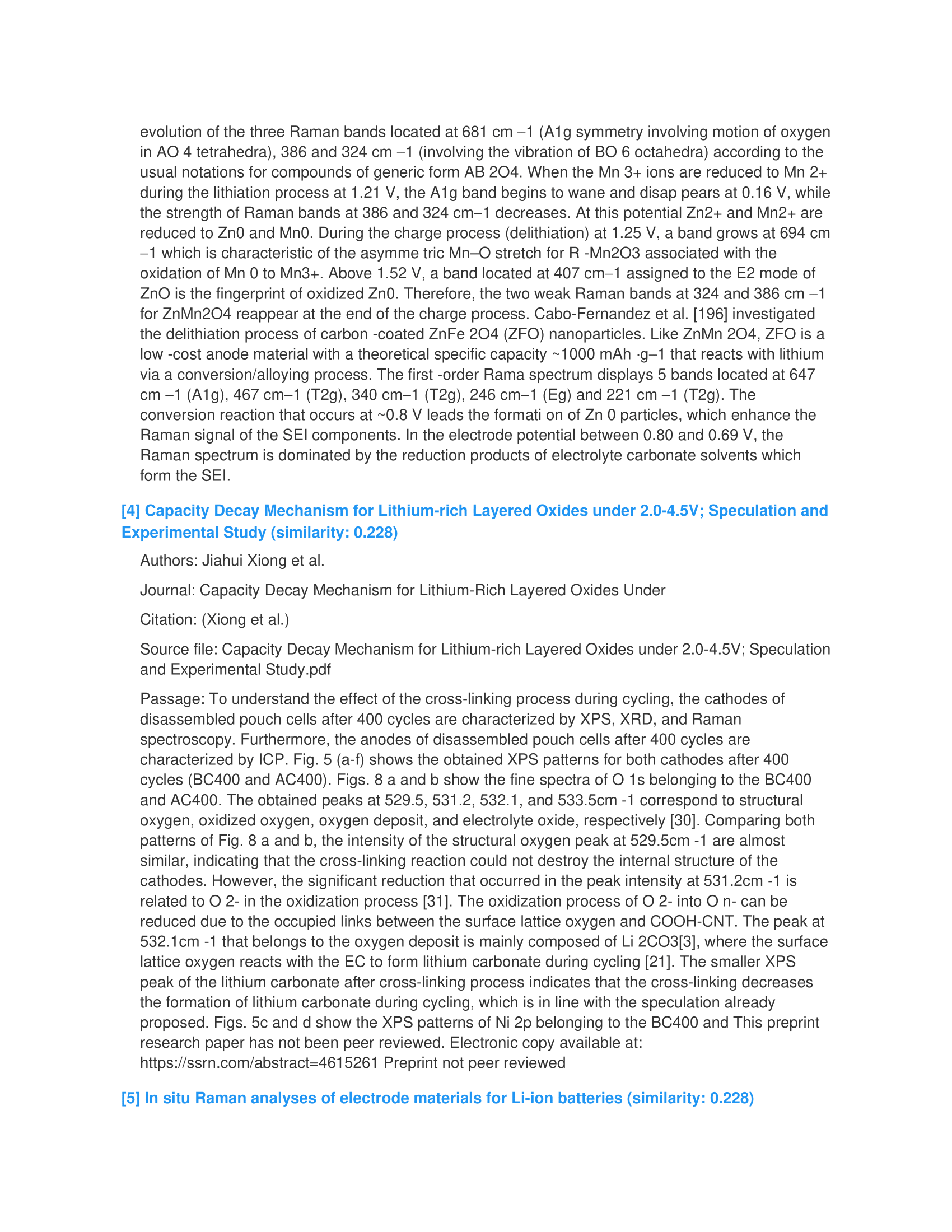}
  \caption{Expert evaluation packet, page 3.}
\end{figure}

\begin{figure}[t]
  \centering
  \includegraphics[width=\linewidth]{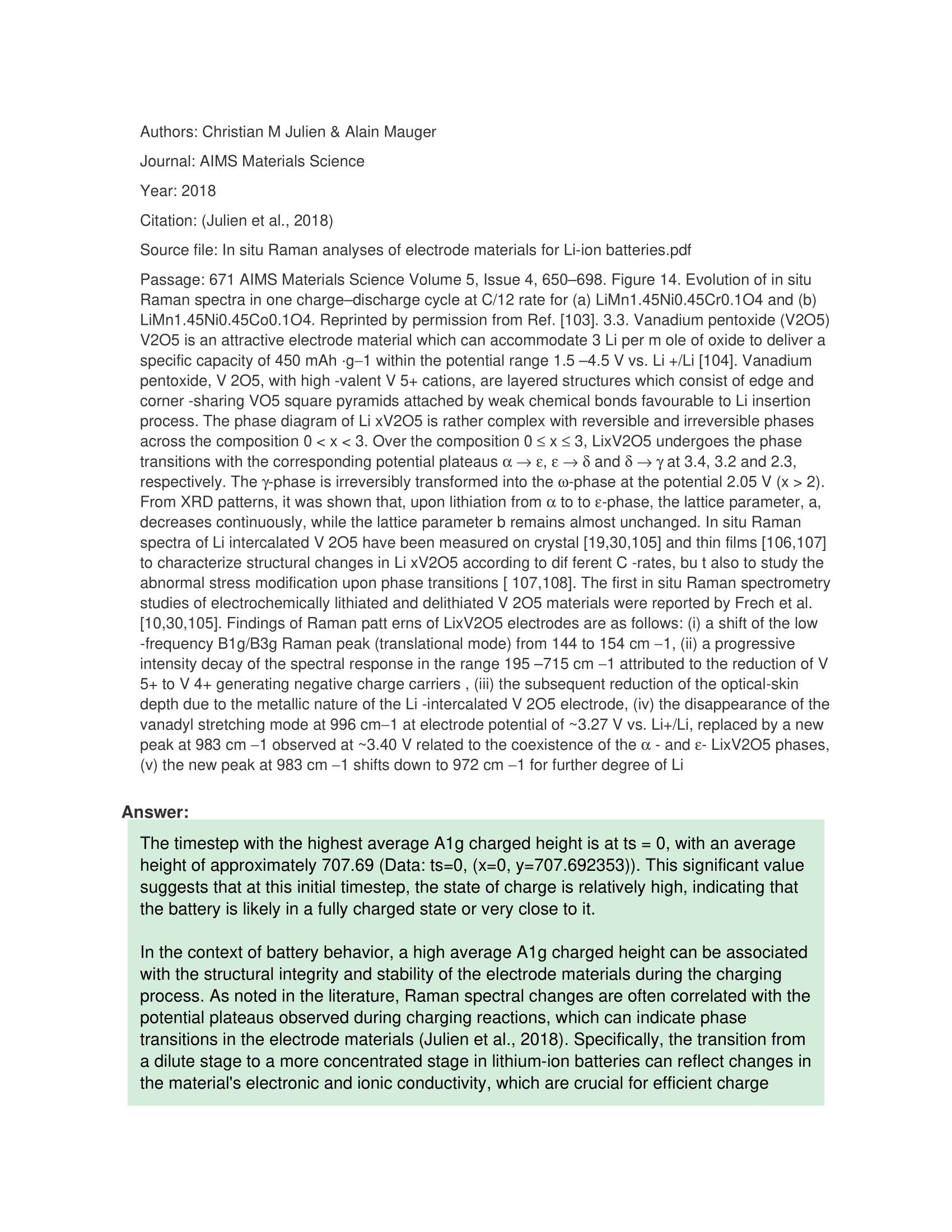}
  \caption{Expert evaluation packet, page 4.}
\end{figure}

\begin{figure}[t]
  \centering
  \includegraphics[width=\linewidth]{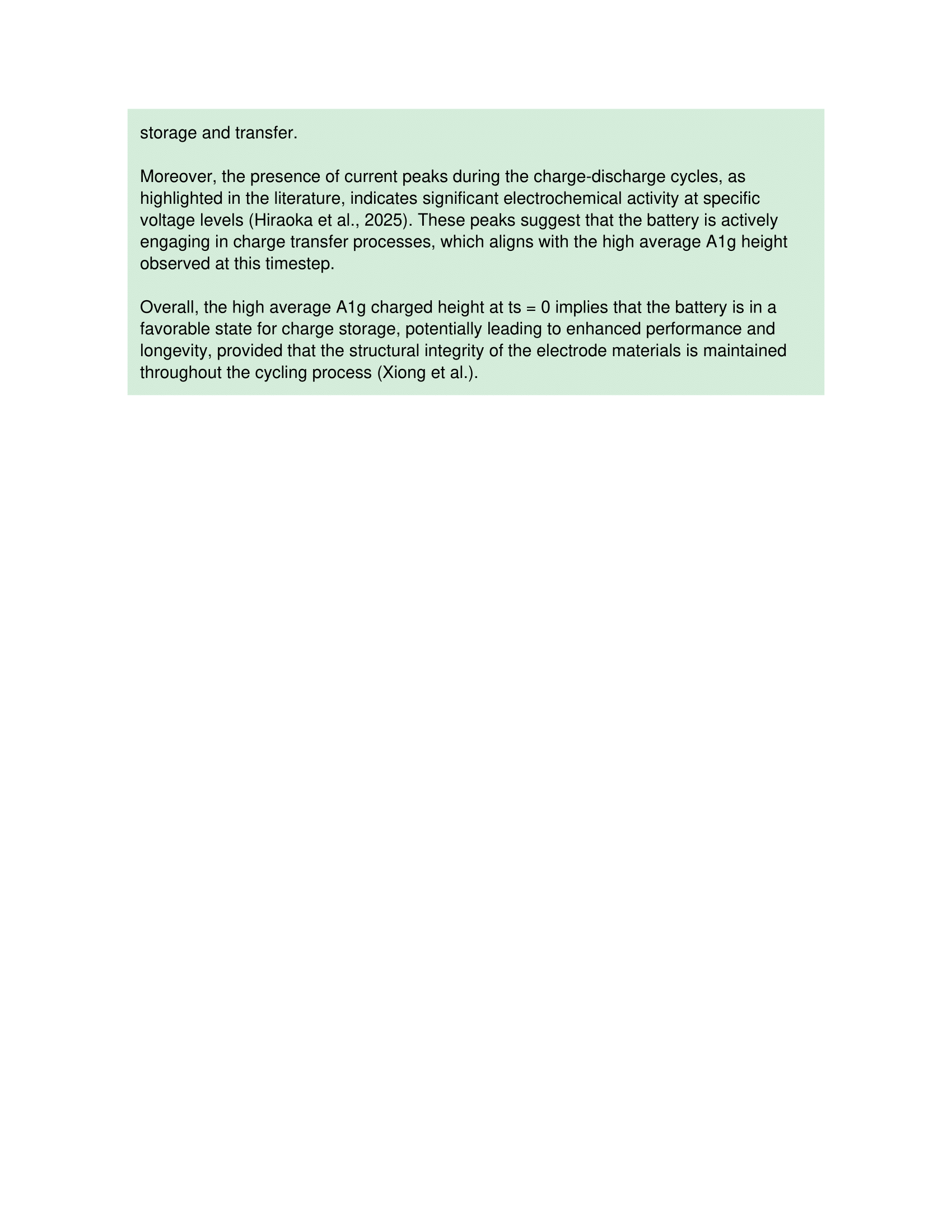}
  \caption{Expert evaluation packet, page 5.}
\end{figure}

\begin{figure}[t]
  \centering
  \includegraphics[width=\linewidth]{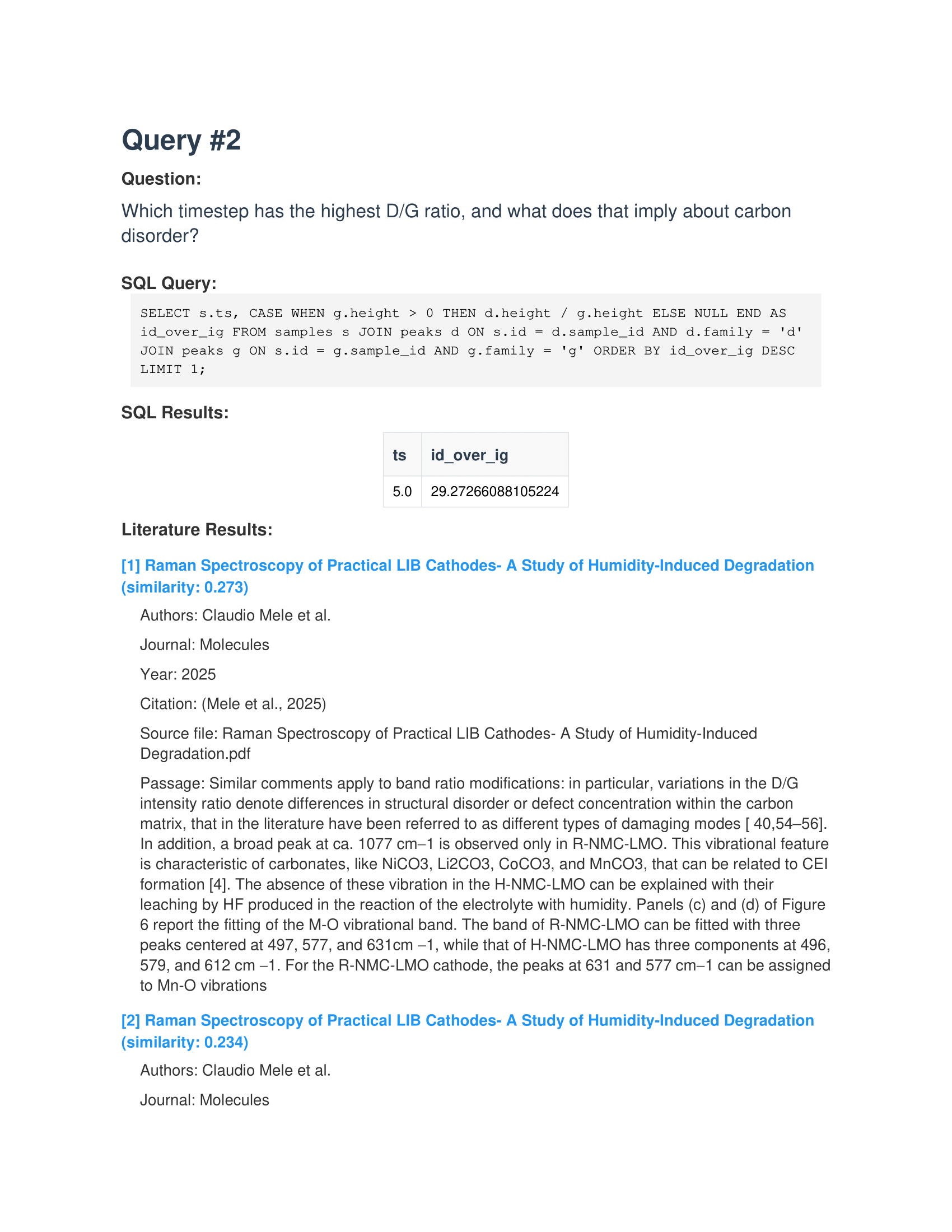}
  \caption{Expert evaluation packet, page 6.}
\end{figure}

\begin{figure}[t]
  \centering
  \includegraphics[width=\linewidth]{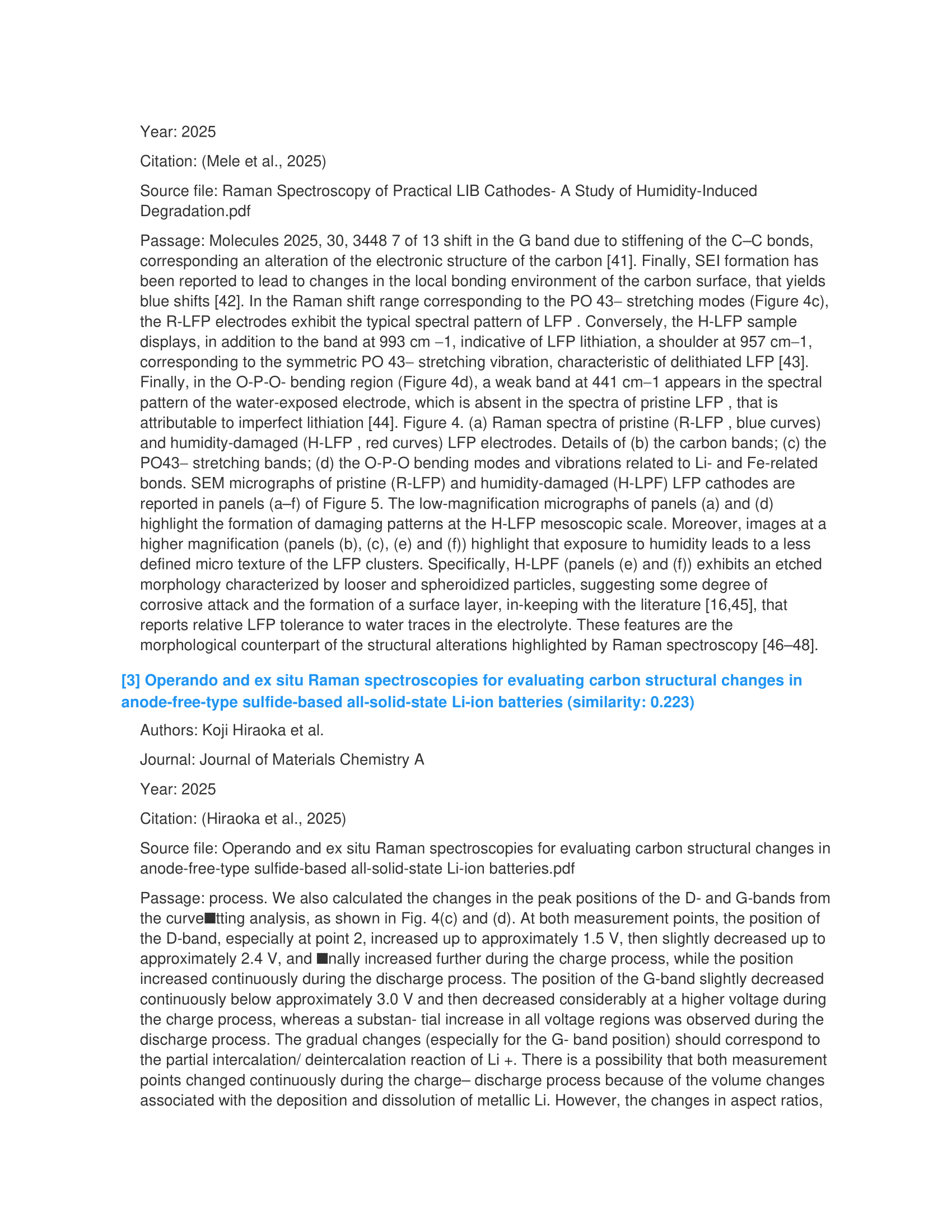}
  \caption{Expert evaluation packet, page 7.}
\end{figure}

\begin{figure}[t]
  \centering
  \includegraphics[width=\linewidth]{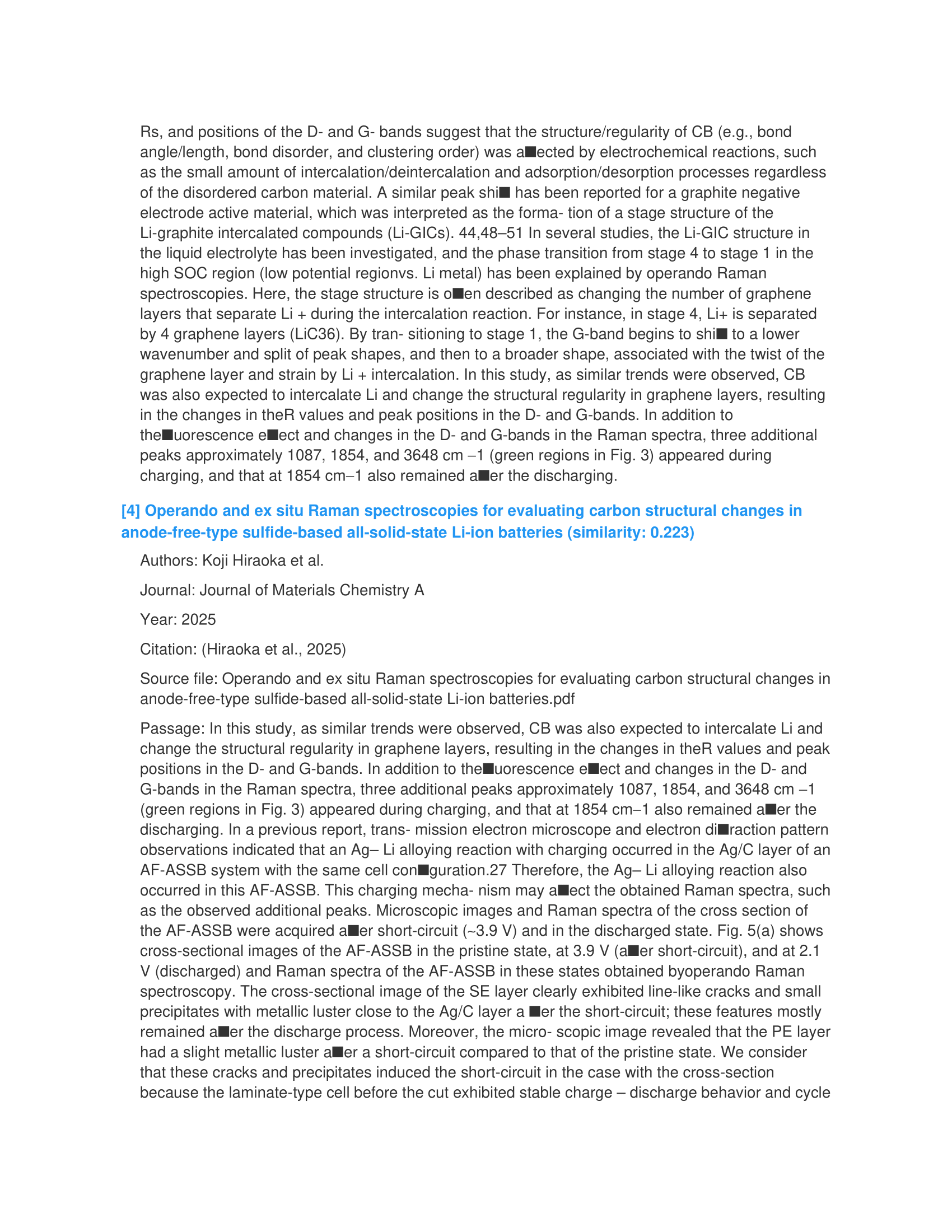}
  \caption{Expert evaluation packet, page 8.}
\end{figure}

\begin{figure}[t]
  \centering
  \includegraphics[width=\linewidth]{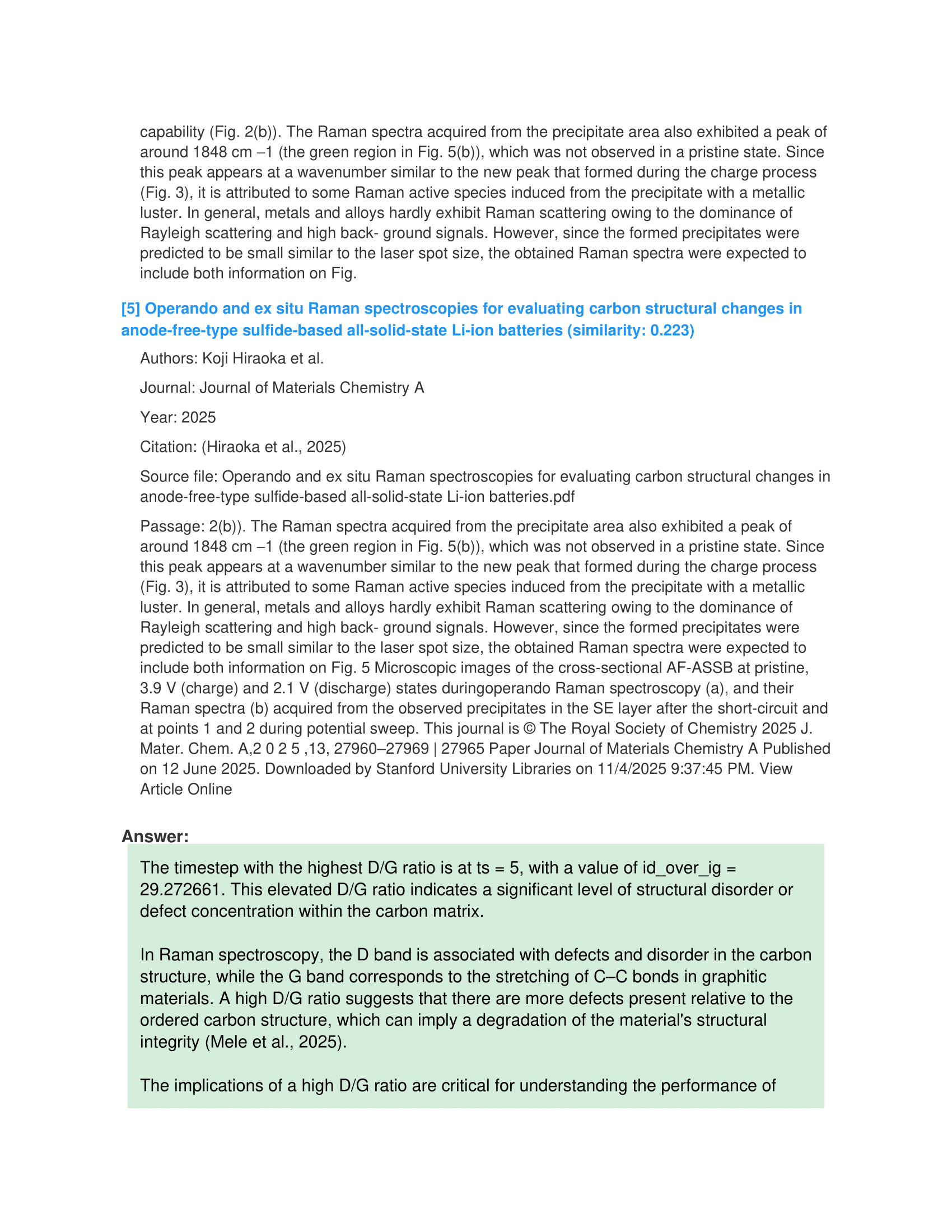}
  \caption{Expert evaluation packet, page 9.}
\end{figure}

\begin{figure}[t]
  \centering
  \includegraphics[width=\linewidth]{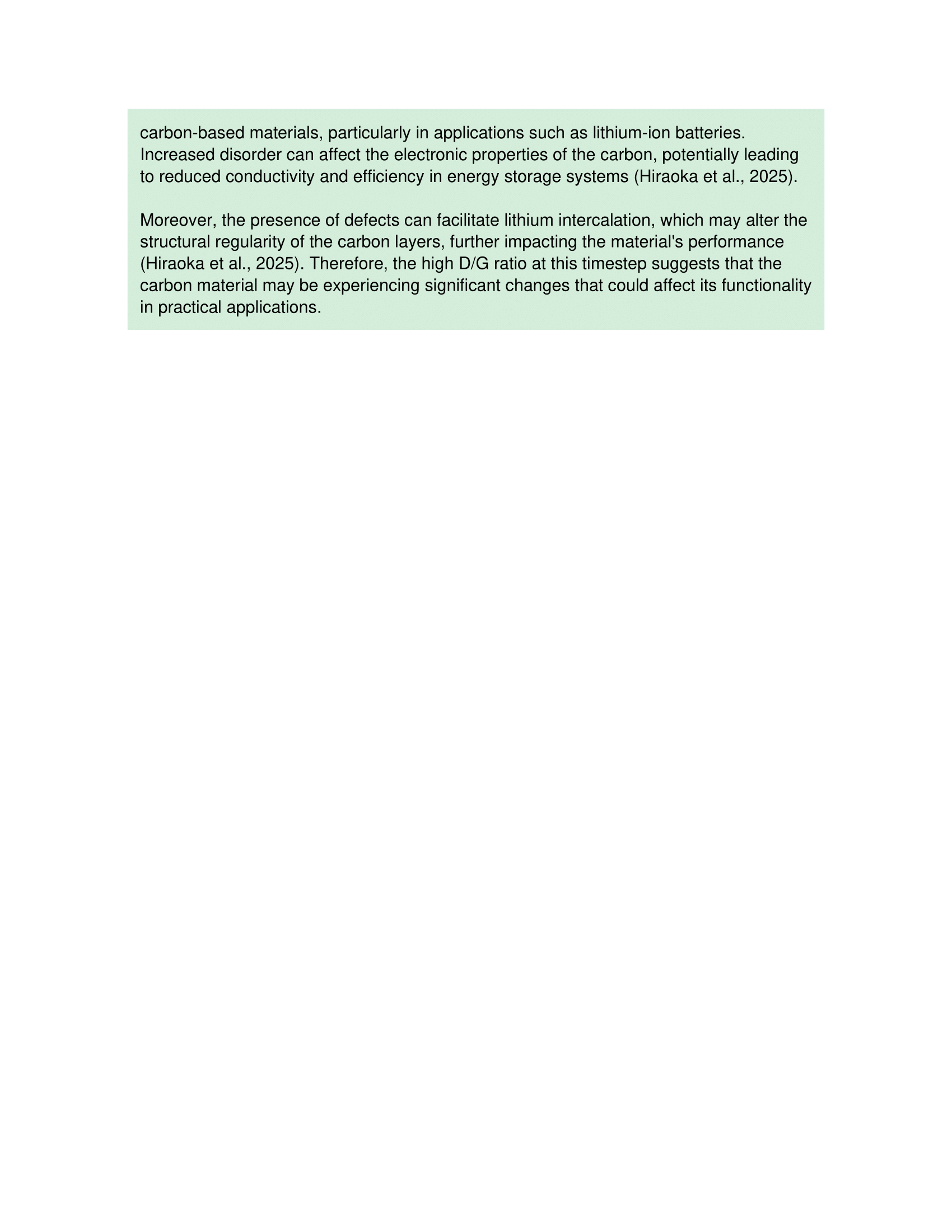}
  \caption{Expert evaluation packet, page 10.}
\end{figure}

\clearpage

\section{Literature Corpus: Full Citation List}
\label{app:corpus-citations}

\begin{itemize}

\item Koji Hiraoka, Yoshiki Yokoyama, Sarina Mine, Kazuo Yamamoto, and Shiro Seki. 2025.
Advanced Raman spectroscopy for battery applications: Materials characterization and operando measurements.
\textit{APL Energy}, 3(2).

\item Jiahui Xiong, Soroosh Mahmoodi, Zhimin Huang, and Shengwen Zhong.
Capacity Decay Mechanism for Lithium-Rich Layered Oxides Under 2.0--4.5V; Speculation and Experimental Study.
\textit{SSRN} (preprint). DOI: 10.2139/ssrn.4631123.

\item Eibar Flores, Petr Nov{\'a}k, and Erik J. Berg. 2018.
In situ and operando Raman spectroscopy of layered transition metal oxides for Li-ion battery cathodes.
\textit{Frontiers in Energy Research}, 6:82.

\item Christian M. Julien and Alain Mauger. 2018.
In situ Raman analyses of electrode materials for Li-ion batteries.
\textit{AIMS Materials Science}, 5(4):650--698.

\item Jinglei Lei, Frank McLarnon, and Robert Kostecki. 2005.
In situ Raman microscopy of individual LiNi$_{0.8}$Co$_{0.15}$Al$_{0.05}$O$_2$ particles in a Li-ion battery composite cathode.
\textit{The Journal of Physical Chemistry B}, 109(2):952--957.

\item M. A. Caba{\~n}ero, Markus Hagen, and E. Quiroga-Gonz{\'a}lez. 2021.
In-operando Raman study of lithium plating on graphite electrodes of lithium ion batteries.
\textit{Electrochimica Acta}, 374:137487.

\item Marco-Tulio Fonseca Rodrigues, Victor A. Maroni, David J. Gosztola, Koffi P. C. Yao, Kaushik Kalaga, Ilya A. Shkrob, and Daniel P. Abraham. 2018.
Lithium acetylide: A spectroscopic marker for lithium deposition during fast charging of Li-ion cells.
\textit{ACS Applied Energy Materials}, 2(1):873--881.

\item Koji Hiraoka, Junichi Sakabe, Naoki Suzuki, and Shiro Seki. 2025.
Operando and ex situ Raman spectroscopies for evaluating carbon structural changes in anode-free-type sulfide-based all-solid-state Li-ion batteries.
\textit{Journal of Materials Chemistry A}.

\item Rose E. Ruther, Andrew F. Callender, Hui Zhou, Surendra K. Martha, and Jagjit Nanda. 2014.
Raman microscopy of lithium-manganese-rich transition metal oxide cathodes.
\textit{Journal of The Electrochemical Society}, 162(1):A98.

\item Claudio Mele, Filippo Ravasio, Andrea Casalegno, Elisa Emanuele, Claudio Rabissi, and Benedetto Bozzini. 2025.
Raman Spectroscopy of Practical LIB Cathodes: A Study of Humidity-Induced Degradation.
\textit{Molecules}, 30(16):3448. DOI: 10.3390/molecules30163448.

\item Eibar Flores, Petr Novak, Ulrich Aschauer, and Erik J. Berg. 2019.
Cation ordering and redox chemistry of layered Ni-rich Li$_x$Ni$_{1-2y}$Co$_y$Mn$_y$O$_2$: an operando Raman spectroscopy study.
\textit{Chemistry of Materials}, 32(1):186--194.

\item Vijay A. Sethuraman, Laurence J. Hardwick, Venkat Srinivasan, and Robert Kostecki. 2010.
Surface structural disordering in graphite upon lithium intercalation/deintercalation.
\textit{Journal of Power Sources}, 195(11):3655--3660. DOI: 10.1016/j.jpowsour.2009.12.034.

\item Ermanno Miele, Wesley M. Dose, Ilya Manyakin, Michael H. Frosz, Zachary Ruff, Michael F. L. De Volder, Clare P. Grey, Jeremy J. Baumberg, and Tijmen G. Euser. 2022.
Hollow-core optical fibre sensors for operando Raman spectroscopy investigation of Li-ion battery liquid electrolytes.
\textit{Nature Communications}, 13(1). DOI: 10.1038/s41467-022-29330-4.

\item Adrian Lindner, Hannes Radinger, Frieder Scheiba, and Helmut Ehrenberg. 2022.
Structure--activity correlation of thermally activated graphite electrodes for vanadium flow batteries.
\textit{RSC Advances}, 12(22):14119--14126. DOI: 10.1039/D2RA02368G.

\item Gozde Oney, Federico Monaco, Saptarshee Mitra, Asma Medjahed, Manfred Burghammer, Dmitry Karpov, Marta Mirolo, Jakub Drnec, Isabelle C. Jolivet, Quentin Arnoux, Samuel Tardif, Quentin Jacquet, and Sandrine Lyonnard. 2025.
Dead, Slow, and Overworked Graphite: Operando X-Ray Microdiffraction Mapping of Aged Electrodes.
\textit{Advanced Energy Materials}, 15(38). DOI: 10.1002/aenm.202502032.

\item Dominika A. Buchberger, Bartosz Hamankiewicz, Monika Michalska, Alicja G{\l}aszczka, and Andrzej Czerwinski. 2024.
Ex Situ Raman Mapping of LiMn$_2$O$_4$ Electrodes Cycled in Lithium-Ion Batteries.
\textit{ACS Omega}, 9(28):30381--30391. DOI: 10.1021/acsomega.4c01480.

\item Jun-Wei Yin, Yi-Meng Wu, Xin-Yu Liu, Jing Li, Peng-Fei Wang, Zong-Lin Liu, Lin-Lin Wang, Jie Shu, and Ting-Feng Yi. 2025.
Insights into degradation mechanisms and engineering strategies of layered manganese-based oxide cathodes for sodium-ion battery.
\textit{Green Energy \& Environment}. DOI: 10.1016/j.gee.2025.07.013.

\item Dominika A. Buchberger, Maciej Boczar, Jacek B. Jasinski, and Andrzej Czerwi\'{n}ski. 2025.
Raman spectroscopy complemented with XRD and TEM for studying structural evolution in initial cycles of LiNi$_{1/3}$Mn$_{1/3}$Co$_{1/3}$O$_2$ cathode material.
\textit{Discover Nano}, 20(1). DOI: 10.1186/s11671-025-04319-2.

\item Marcel Heber, Kathrin Hofmann, and Christian Hess. 2022.
Raman Diagnostics of Cathode Materials for Li-Ion Batteries Using Multi-Wavelength Excitation.
\textit{Batteries}, 8(2):10. DOI: 10.3390/batteries8020010.

\item Ruichuan Yuan, Yiwen Guo, Ilke Gurgan, Nahian Siddique, Yu-Sheng Li, Seokhoon Jang, Gina A. Noh, and Seong H. Kim. 2025.
Raman spectroscopy analysis of disordered and amorphous carbon materials: A review of empirical correlations.
\textit{Carbon}, 238:120214. DOI: 10.1016/j.carbon.2025.120214.

\item Tianxun Cai, Mingzhi Cai, Jinxiao Mu, Siwei Zhao, Hui Bi, Wei Zhao, Wujie Dong, and Fuqiang Huang. 2023.
High-Entropy Layered Oxide Cathode Enabling High-Rate for Solid-State Sodium-Ion Batteries.
\textit{Nano-Micro Letters}, 16(1). DOI: 10.1007/s40820-023-01232-0.

\item Xinyu Liu, Jaehoon Choi, Zhen Xu, Clare P. Grey, Simon Fleischmann, and Alexander C. Forse. 2024.
Raman Spectroscopy Measurements Support Disorder-Driven Capacitance in Nanoporous Carbons.
\textit{Journal of the American Chemical Society}, 146(45):30748--30752. DOI: 10.1021/jacs.4c10214.

\item Sven Jovanovic, Peter Jakes, Steffen Merz, R\"{u}diger-A. Eichel, and Josef Granwehr. 2021.
Lithium intercalation into graphite: In operando analysis of Raman signal widths.
\textit{Electrochemical Science Advances}, 2(4). DOI: 10.1002/elsa.202100068.

\item M. J. Madito. 2025.
Revisiting the Raman disorder band in graphene-based materials: A critical review.
\textit{Vibrational Spectroscopy}, 139:103814. DOI: 10.1016/j.vibspec.2025.103814.

\item Debbie Zhuang and Martin Z. Bazant. 2022.
Theory of Layered-Oxide Cathode Degradation in Li-ion Batteries by Oxidation-Induced Cation Disorder.
\textit{Journal of The Electrochemical Society}, 169(10):100536. DOI: 10.1149/1945-7111/ac9a09.

\item Yasutaka Matsuda, Naoaki Kuwata, Tatsunori Okawa, Arunkumar Dorai, Osamu Kamishima, and Junichi Kawamura. 2019.
In situ Raman spectroscopy of LiCoO$_2$ cathode in Li/Li$_3$PO$_4$/LiCoO$_2$ all-solid-state thin-film lithium battery.
\textit{Solid State Ionics}, 335:7--14. DOI: 10.1016/j.ssi.2019.02.010.

\item Lukas Karapin-Springorum, Asia Sarycheva, Andrew Dopilka, Hyungyeon Cha, Muhammad Ihsan-Ul-Haq, Jonathan M. Larson, and Robert Kostecki. 2025.
An infrared, Raman, and X-ray database of battery interphase components.
\textit{Scientific Data}, 12(1). DOI: 10.1038/s41597-024-04236-6.

\item R. Hausbrand, G. Cherkashinin, H. Ehrenberg, M. Gr\"{o}ting, M. Albe, C. Hess, and W. Jaegermann. 2015.
Fundamental degradation mechanisms of layered oxide Li-ion battery cathode materials: Methodology, insights and novel approaches.
\textit{Materials Science and Engineering: B}, 192:3--25. DOI: 10.1016/j.mseb.2014.11.014.

\item Slaheddine Jabri, Luciana Pitta Bauermann, and Matthias Vetter. 2023.
Raman spectrometry measurements for the 2D mapping of the degradation products on aged graphite anodes of cylindrical Li-ion battery cells.
\textit{AIP Advances}, 13(11). DOI: 10.1063/5.0171158.

\item Ray H. Baughman, Anvar A. Zakhidov, and Walt A. de Heer. 2007.
Charge transfer in carbon nanotube actuators investigated using in situ Raman spectroscopy.
\textit{Advanced Materials}, 19(19):3213--3218. DOI: 10.1002/adma.200602660.

\item Yong Soo Cho, Younghwa Kim, Jaejin Kim, and others. 2019.
Detection of secondary phase in NMC811(OH)$_2$ precursor using X-ray powder diffraction and Raman spectroscopy.
\textit{Ceramics International}, 45(18):24036--24042. DOI: 10.1016/j.ceramint.2019.08.246.

\item Robert J. Nemanich and S. A. Solin. 1979.
Distinguishing disorder-induced bands from allowed Raman bands in graphite.
\textit{Physical Review B}, 20(2):392--401. DOI: 10.1103/PhysRevB.20.392.

\item Rodolfo P. Vidano and David B. Fischbach. 1981.
Fundamentals, overtones, and combinations in the Raman spectrum of graphite.
\textit{Solid State Communications}, 39(2):341--344. DOI: 10.1016/0038-1098(81)90686-4.

\item Hannes Radinger, Martin Petz, Helmut Ehrenberg, and others. 2022.
Hierarchical structuring of NMC111-cathode materials in lithium-ion batteries: An in-depth study on the influence of primary and secondary particle sizes on electrochemical performance.
\textit{Journal of Power Sources}, 528:231210. DOI: 10.1016/j.jpowsour.2022.231210.

\item Yihan Zhu, Shanthi Murali, Wei Cai, and others. 2017.
In situ electrochemical Raman investigation of charge storage in rGO and N-doped rGO.
\textit{Carbon}, 123:424--432. DOI: 10.1016/j.carbon.2017.07.065.

\item Arne Sadezky, Heinrich Muckenhuber, Heiko Grothe, Reinhard Niessner, and Ulrich P{\"o}schl. 2005.
Raman spectroscopic investigations of activated carbon materials.
\textit{Carbon}, 43(8):1731--1742. DOI: 10.1016/j.carbon.2005.02.018.

\item Naoaki Kuwata, Junichi Kawamura, and others. 2017.
Raman spectroscopy for LiNi$_{1/3}$Mn$_{1/3}$Co$_{1/3}$O$_2$ composite positive electrodes in all-solid-state lithium batteries.
\textit{Solid State Ionics}, 304:71--76. DOI: 10.1016/j.ssi.2016.11.016.

\item Mildred S. Dresselhaus, Ado Jorio, and Riichiro Saito. 2005.
Raman spectroscopy of carbon materials: structural basis of observed spectra.
\textit{Physics Reports}, 409(2):47--99. DOI: 10.1016/j.physrep.2004.10.006.

\item Xing Zhang and others. 2020.
Surface changes of LiNi$_x$Mn$_y$Co$_{1-x-y}$O$_2$ in Li-ion batteries using in situ surface-enhanced Raman spectroscopy.
\textit{Journal of Materials Chemistry A}, 8(9):4560--4569. DOI: 10.1039/C9TA13634A.

\item Wei Li and others. 2017.
The carbon-based 3D-hierarchical cathode architecture for Li-ion batteries.
\textit{Advanced Functional Materials}, 27(15):1606439. DOI: 10.1002/adfm.201606439.

\item Steven Sloop and others. 2021.
Operando Raman spectroscopic analysis for electrolyte/electrode interface reaction in lithium--sulfur batteries with sparingly solvating electrolyte.
\textit{Journal of Power Sources}, 507:230282. DOI: 10.1016/j.jpowsour.2021.230282.

\item Bruno Scrosati and others. 2013.
Exploring carbon electrode parameters in Li--O$_2$ cells: Li$_2$O$_2$ and Li$_2$CO$_3$ formation.
\textit{Energy \& Environmental Science}, 6(5):1440--1445. DOI: 10.1039/C3EE24284G.

\item Sungho Park and others. 2019.
Characterization of pitch carbon coating properties affecting the electrochemical behavior of silicon nanoparticle lithium-ion battery anodes.
\textit{Electrochimica Acta}, 320:134557. DOI: 10.1016/j.electacta.2019.134557.

\item Masashi Inaba and others. 2016.
In situ surface-enhanced Raman spectroelectrochemistry reveals the molecular conformation of electrolyte additives in Li-ion batteries.
\textit{Journal of Power Sources}, 307:504--512. DOI: 10.1016/j.jpowsour.2015.12.060.

\item M. Matsui and others. 2019.
In operando Raman microscopy of the Cu/Li$_{1.5}$Al$_{0.5}$Ge$_{1.5}$(PO$_4$)$_3$ solid electrolyte interphase.
\textit{Journal of Materials Chemistry A}, 7(27):16157--16166. DOI: 10.1039/C9TA03689G.

\item Markus Hagen and others. 2020.
Operando Raman spectroscopy for investigating lithium deposition/dissolution and diffusion at the microelectrode surface.
\textit{Electrochimica Acta}, 354:136744. DOI: 10.1016/j.electacta.2020.136744.

\item Yutaka Yamada and others. 2021.
In operando Raman spectroscopy reveals Li-ion solvation in lithium metal batteries.
\textit{Nature Communications}, 12(1):5403. DOI: 10.1038/s41467-021-25669-2.

\item Walter Van Schalkwijk and Bruno Scrosati (editors). 2002.
\textit{Materials for Lithium-Ion Batteries}.
Springer, Boston. DOI: 10.1007/978-1-4419-9269-3.

\item Victor Stancovski and Simona Badilescu. 2014.
In situ Raman spectroscopic--electrochemical studies of lithium-ion battery materials: a historical overview.
\textit{Journal of Applied Electrochemistry}, 44(1):23--43. DOI: 10.1007/s10800-013-0628-0.

\item Chitturi Venkateswara Rao and others. 2014.
Investigations on electrochemical behavior and structural stability of Li$_{1.2}$Mn$_{0.54}$Ni$_{0.13}$Co$_{0.13}$O$_2$ lithium-ion cathodes via in situ and ex situ Raman spectroscopy.
\textit{The Journal of Physical Chemistry C}, 118(26):14133--14141. DOI: 10.1021/jp501777v.

\end{itemize}
\clearpage

\section{Expert Evaluation Form}
\label{app:expert-form}

Expert reviewers evaluated each SpectraQuery response using an ordinal 5-point Likert scale, where
\textbf{1 = Strongly Disagree / Poor} and \textbf{5 = Strongly Agree / Excellent}. For each query--answer pair, reviewers provided ratings for the following criteria, along with an optional free-response comment.

\begin{enumerate}
  \item \textbf{Scientific Accuracy:} ``The response is scientifically correct and free of factual or interpretive errors.''
  \item \textbf{Grounding in Evidence:} ``The response is well-supported by both Raman data and cited literature excerpts.''
  \item \textbf{Relevance to the Question:} ``The response directly answers the specific scientific question posed.''
  \item \textbf{Clarity and Readability:} ``The explanation is clearly written, logically organized, and easy to follow.''
  \item \textbf{Depth and Insight:} ``The response demonstrates meaningful scientific reasoning or new insight beyond surface-level description.''
  \item \textbf{Completeness:} ``The response includes all key aspects or considerations relevant to the question.''
  \item \textbf{Interpretability of Citations / References:} ``I can easily trace the cited figures, tables, or literature passages to verify claims.''
  \item \textbf{Usefulness for Expert Workflow:} ``This output would meaningfully aid my own data analysis or interpretation.''
  \item \textbf{Optional free response:} What did you like or dislike about this answer?
\end{enumerate}

\end{document}